\documentclass{article} 
\usepackage[final]{colm2026_conference}

\usepackage{microtype}
\usepackage{url}
\usepackage{booktabs}

\usepackage{titletoc}
\usepackage{amsmath}
\usepackage{amssymb}
\usepackage{mathtools}
\usepackage{amsthm}

\usepackage{subcaption}
\usepackage{wrapfig}
\usepackage[table]{xcolor}
\definecolor{CS}{HTML}{E6F0FA}
\definecolor{MCQ}{HTML}{EAF7E6}
\definecolor{QA}{HTML}{FFF3E6}
\definecolor{PARA}{HTML}{F0E6FA}
\definecolor{TFQ}{HTML}{F9E6EA}
\definecolor{mygray}{HTML}{808080}
\definecolor{mylightgray}{HTML}{F2F2F2}
\definecolor{customgreen}{HTML}{2ECC40}
\usepackage{hyperref}

\usepackage[normalem]{ulem}

\usepackage{url}
\usepackage{enumitem}  
\usepackage{tcolorbox}
\usepackage{wrapfig}
\usepackage{multirow}
\usepackage[table]{xcolor}
\usepackage{xspace}
\usepackage{soul}
\usepackage{xcolor}
\newcommand{\dataset}{\textsc{Mixture}\xspace}
\newcommand{\method}{\textit{LM-Mixup}\xspace}
\newcommand{\task}{\emph{Instruction Distillation}\xspace}

\usepackage{lineno}

\definecolor{darkblue}{rgb}{0, 0, 0.5}
\hypersetup{colorlinks=true, citecolor=darkblue, linkcolor=darkblue, urlcolor=darkblue}

\title{LM-mixup: Text Data Augmentation via \\Language Model based Mixup}

\author{Zhijie Deng$^1$~~~~Zhouan Shen$^1$ ~~~~Ling Li$^1$ ~~~~Yao Zhou ~~~~ \textbf{Zhaowei Zhu}$^{2,3}$  \\\textbf{Yanji He}$^1$ ~~~~\textbf{Wei Wang}* $^1$ ~~~~\textbf{Jiaheng Wei\thanks{Corresponding to jiahengwei@hkust-gz.edu.cn, weiwcs@hkust-gz.edu.cn.}}~~$^{1,3}$ \vspace{0.8ex} \\
$^1$The Hong Kong University of Science and Technology (Guangzhou)\\$^2$BIAI, Zhejiang University of Technology~~~~$^3$D5Data.ai\\ 
\texttt{zdeng190@connect.hkust-gz.edu.cn},~~\texttt{jiahengwei@hkust-gz.edu.cn}\\
}

\begin{document}

\ifcolmsubmission
\linenumbers
\fi

\maketitle

\begin{abstract}
Instruction tuning is crucial for aligning Large Language Models (LLMs), while the quality of instruction-following data varies significantly. In addition, abundant low-quality data is frequently discarded, leading to substantial information loss. Existing data augmentation methods struggle to augment these low-quality data effectively, and the evaluation of such techniques remains poorly defined. To address this, we formally define the task of \emph{Instruction Distillation}: distilling multiple low-quality and redundant inputs into high-quality and coherent instruction-output pairs. Specifically, we introduce a comprehensive data construction pipeline to create \dataset, a 144K-sample dataset pairing low-quality or semantically redundant imperfect instruction clusters with their high-quality distillations. We then introduce \emph{LM-Mixup}, first performing supervised fine-tuning on \dataset and then optimizing it with reinforcement learning. This process uses three complementary reward signals: quality, semantic alignment, and format compliance, via Group Relative Policy Optimization (GRPO). We demonstrate that \emph{LM-Mixup} effectively augments imperfect datasets: fine-tuning LLMs on its distilled data, which account for only about 3\% of the entire dataset, not only surpasses full-dataset training, but also competes with state-of-the-art high-quality data selection methods across multiple benchmarks. Our work establishes that low-quality data are a valuable resource when properly distilled and augmented with \emph{LM-Mixup}, significantly enhancing the efficiency and performance of instruction-tuned LLMs. Our code and data are publicly available at \url{https://github.com/yuu250/LM-mixup}.
\end{abstract}

\section{Introduction}

In recent years, large language models (LLMs) have achieved notable progress in natural language processing and multimodal understanding~\citep{team2023gemini,singhal2023large,deng2025guard,li2024georeasoner,li2025recognition}. This progress stems not only from improved architectures and larger scales but also from more efficient ways for models to learn and apply knowledge~\citep{patil2025advancing,dredze2025amuro}. While the conventional view holds that high-quality human alignment requires massive annotated data~\citep{kim2024aligning,kopf2023openassistant}, recent studies show that LLMs acquire most knowledge during pre-training~\citep{brown2020language,roberts2020much}. Only a small, carefully curated dataset is sufficient for effective alignment in instruction tuning or supervised fine-tuning (SFT)~\citep{he2024shed,wei2023instructiongpt}, so many works now focus on selecting high-quality data, demonstrating that fine-tuning on such subsets alone can already yield strong performance~\citep{pang2024improving,fu2025t,jha2023limit}. This shifts the research focus from “more data” to “better data”, emphasizing efficient high-quality data selection for model improvement.

\begin{figure}
    \centering
    \includegraphics[width=\linewidth]{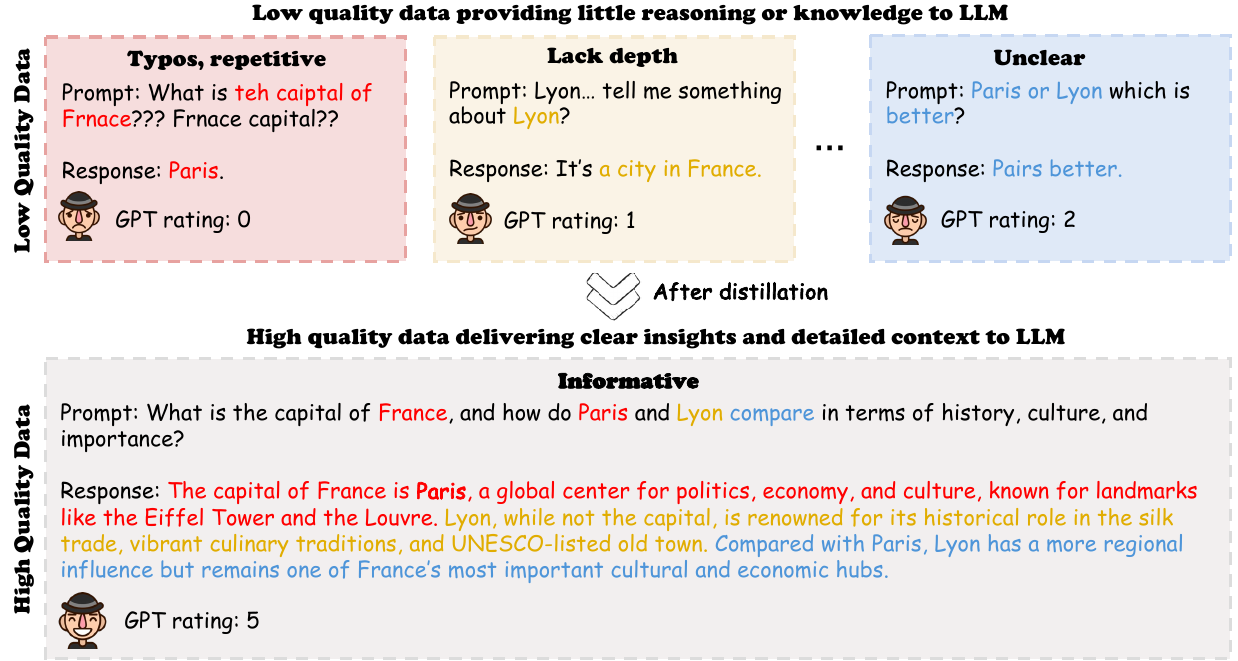}
    \caption{The goal of \task. Low-quality samples show issues such as typos, shallow content, and unclear intent, each receiving low GPT ratings. After distillation, they are fused into a single high-quality sample with clear, informative, and context-rich content. Ratings are on a 5-point scale.  Additional case studies are in Appendix~\ref{sec:case_study}.}
    \label{fig:quality_examples}
\end{figure}

However, high-quality samples are scarce and costly, while real-world datasets contain abundant redundant or low-quality data, leading to significant information waste. This gap mainly arises from data characteristics: low-quality samples are often simple or repetitive with limited learning signals, while high-quality samples involve complex reasoning or rich knowledge, making them more beneficial for training~\citep{morishita2024enhancing}, as shown in Figure~\ref{fig:quality_examples}. Moreover, in many specialized domains or low-resource settings (e.g., low-resource machine translation and domain-specific tasks such as law or medicine), the scarcity of high-quality data is widely regarded as a key bottleneck that limits further progress~\citep{low1,low2,low3}.
Recently, some works have begun exploring ways to enhance low-quality data to unlock their potential; however, most existing approaches still rely on heuristic rules or handcrafted templates, struggling to substantially enrich their information content or complexity~\citep{chai2025text,zhu2025tag,lee2024llm2llm}. This raises a key question: \textbf{\textit{can we fully exploit low-quality data and transform it into a valuable resource for improving LLM training?}}

In this work, we study how to efficiently leverage low-quality data and introduce the \task paradigm: given topic-related but sparse and incomplete inputs, the goal is to aggregate and rewrite them into a single information-dense target.
To facilitate this paradigm, we construct \dataset, a Wikipedia-based dataset with about 144K instances across five task types, providing hierarchical mappings from multiple low-quality inputs to a single high-quality output, as shown in Figure \ref{fig:pipeline}. Each high quality data pair comes with 2 to 20 controlled low quality variants and optional chain-of-thought supervision. To further improve diversity in the dataset and robustness during the training process, cross-topic mixing and noise injection are added.

Since SFT concentrates on memorizing answers~\citep{li2025preservingdiversitysupervisedfinetuning,chu2025sftmemorizesrlgeneralizes} and fails to explore diverse strategies for distilling low-quality samples into high-quality outputs, we adopt GRPO~\citep{guo2025deepseek} to optimize the generation process. Building on \dataset, we further train \method with GRPO to fully exploit its potential. 
Concretely, we perform cold-start pre-training on a subset of \dataset to equip the model with the basic ability to generate high-quality outputs; then, we apply GRPO-based reinforcement learning to jointly optimize output quality along three dimensions: quality, semantic alignment, and format compliance. 
\method significantly outperforms SFT and selective baselines in multiple tasks, with small-scale models even surpassing strong instruction models under direct prompting; besides, a small amount of original high-quality data combined with distilled results from low-quality data (totally 10K) can achieve or exceed the performance of large-scale datasets (300K) and advanced data selection methods, demonstrating excellent data efficiency and generalization.

Our contributions are summarized as follows:
\begin{itemize}[leftmargin=*, topsep=-5pt, itemsep=-2pt]
\item We introduce the \textbf{\task task}, which aims to transform low-quality inputs (i.e., sparse, incomplete, etc) into a single information-dense output; to support this paradigm, we construct \dataset, a \textbf{144K-instance Wikipedia-based dataset} with hierarchical mappings from multiple low-quality variants to high-quality targets.
\item We introduce \method, initialized through cold-start pretraining and optimized with GRPO-based reinforcement learning using \textbf{multi-dimensional rewards} (quality, semantic alignment, and format compliance), achieving \textbf{superior performance on the \dataset test set} compared to SFT and strong baselines.
\item Experiments show that training downstream models on the distilled data together with the original high-quality data (only \textbf{$\approx$3\% of the full dataset}), matches or surpasses full-dataset training and advanced data selection methods on public benchmarks, demonstrating the \textbf{significant value of \method enhanced low-quality data}.
\end{itemize}

\section{Related Work}
\label{sec:related_work_append}

\textbf{Data-centric AI.}
Data-centric AI aims to improve model performance by improving the quality of training data rather than only scaling models \citep{Ng2021DataCentric,Ng2022Competition,jiaheng3_mingjie,jiaheng1_vlm,jiaheng4}. 
Existing efforts cover several directions, including data cleaning and error correction \citep{geerts2013llunatic,krishnan2016activeclean,krishnan2017boostclean,rekatsinas2017holoclean,mavrogiorgos2022automated}, data augmentation \citep{zhang2017mixup,jiang2019polar,jiaheng2_zile}, and human-assisted annotation or relabeling for constructing high-quality datasets \citep{wang2018glue,rajpurkar2016squad}. 
More recently, data quality assessment has attracted increasing attention, where methods such as DS2 improve the reliability of LLM-based ratings for data curation \citep{pang2025improvingdataefficiencycurating}. 
Related studies also investigate how to detect and mitigate noisy or mislabeled data, for example through LLM-based correction \citep{wang2024noisegpt} or optimization-based identification of noisy instances \citep{yang2024searching}. 
In contrast to prior work that mainly filters, repairs, or scores existing data, our work studies how to transform multiple low-quality samples into a single higher-quality instance, targeting a more direct utilization of abundant weak-quality data.

\textbf{Mixup methods.}
Mixup was originally proposed as a linear interpolation strategy for data augmentation and has shown strong benefits for robustness and generalization \citep{zhang2017mixup,cao2024survey}. 
Subsequent work extended this idea to text, exploring interpolation at the word, sentence, embedding, and latent representation levels \citep{guo2019augmenting,guo2020nonlinear,sun2020mixup,chen2020mixtext,jindal2020augmenting,zhang2021mixup,chen2022doublemix,kong2022dropmix,zhang2022treemix}. 
Recent methods further improve textual mixup through structural constraints, input-level perturbation, or representation-level fusion \citep{yoon2021ssmix,zheng2023self,yang2024amplify}. 
However, most existing mixup approaches focus on interpolation-based augmentation between training samples. 
Our work differs in both objective and setting: instead of interpolating examples in feature space, we leverage the mixup intuition to fuse multiple low-quality, same-topic inputs into a single higher-quality output, making mixup a tool for data refinement rather than conventional augmentation. 
This distinction is also different from recent instruction merging methods such as MergeIT \citep{cai2025mergeit}, which mainly merge similar medium-quality instructions, whereas we focus on distilling useful signals from noisier low-quality data.

\section{\dataset Dataset}
\label{sec:dataset_pipeline}
\begin{figure}
    \vspace{-9mm}
    \centering
    \includegraphics[width=\linewidth]{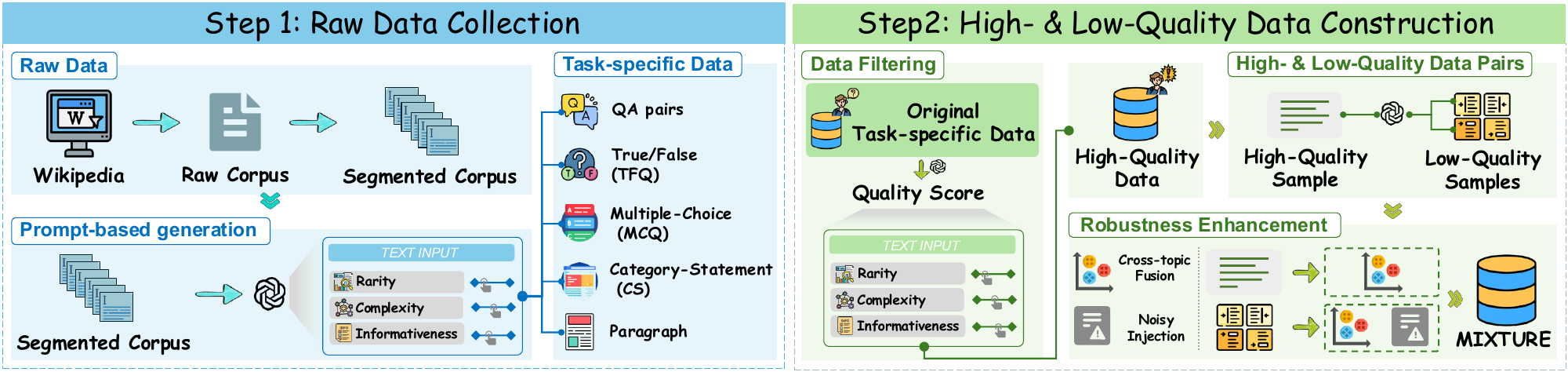}
    \caption{Overview of the \dataset construction pipeline. The process consists of two stages: (1) Raw data collection from Wikipedia, including segmentation and prompt-based generation of five task types; (2) High- and low-quality data construction via LLM-based scoring, multi-variant degradation, cross-topic fusion, and noisy injection, producing the final dataset for training.
    }
    \label{fig:pipeline}
    \vspace{-3mm}
\end{figure}

\subsection{Task Formulation}
The task of \emph{Instruction Distillation} aims to distill multiple potentially imperfect inputs (e.g., redundant or low-quality data) into a single, high-quality instruction--response pair. 
Let $\mathcal{X}$ denote the space of low-quality samples and $\mathcal{Y}$ the set of high-quality texts.
Each training instance consists of a multi-source input
$X=\{\ell_1,\ldots,\ell_k\}\subset\mathcal{X}$ describing the same topic/task
and a reference high-quality target $Y\in\mathcal{Y}$ that is a single high-quality instruction--response pair drawn from the same task.
We denote the training corpus as $\mathcal{D}=\{(X_i,Y_i)\}_{i=1}^n$.
The objective is to learn a generator $f_\theta$ that fuses the multi-source inputs and produces a high-quality text $\hat{Y}=f_\theta(X)$ which (i) preserves the salient information in $X$ while denoising conflicts and increasing information density, (ii) aligns semantically with $X$, and (iii) adheres to the task-specific format.
multiple valid fusions may exist for the same $X$, requiring the model to capture output diversity while maintaining quality.

\subsection{Dataset Statistics}
We introduce \dataset, a dataset specifically designed for \task, comprising five task types: QA pairs, True/False (TFQ), Paragraph, Multiple-Choice Question (MCQ), and Category-Statement (CS). The overall pipeline is illustrated in Figure~\ref{fig:pipeline}.
Overall, \dataset comprises 144,884 samples spanning these five task types, with a balanced distribution across normal, cross-topic, and noisy variants, as shown in Appendix~\ref{ap:data_stat}. 

\subsection{Raw Data Collection}

\textbf{Source Selection.} As shown in Step~1 of Figure~\ref{fig:pipeline}, we use the Wikipedia dataset\footnote{\url{https://huggingface.co/datasets/lucadiliello/english_wikipedia}} as the initial source and sample entries across diverse categories, including domains such as biology, law, and other long-tail areas, to build a broadly covered corpus. To improve quality, we remove overly short entries, extract only plain text, and apply basic deduplication to eliminate redundant content. After filtering, about 10,000 Wikipedia entries remain.

\textbf{Paragraph Segmentation.} Since Wikipedia articles are often long, directly feeding them into LLMs may cause inefficiency and instability~\citep{liu2023lost}. We segment each article into semantically coherent blocks by first splitting into sentences and then greedily concatenating them until a token limit of 512\footnote{We use \texttt{cl100k\_base} tokenizer.} is reached. To balance coherence and boundary effects, we allow optional overlaps and merge very short segments, while over-length sentences are further split at punctuation marks. The final blocks preserve the original order to ensure narrative consistency and traceability. More details can be found in Appendix~\ref{sec:add_details_mixture}.

\begin{figure}
    \centering
    \includegraphics[width=\linewidth]{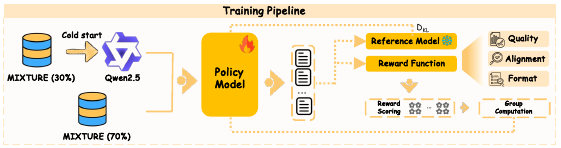}
    \caption{Overview of the training pipeline. The process involves cold-start pretraining on a subset of \dataset, followed by policy optimization using multi-dimensional rewards.}
    \label{fig:training_pipeline}
\end{figure}

\subsection{High- and Low-Quality Data Construction}
\label{sec:high_qual_gen}

\textbf{High-Quality Sample Generation.}
As shown in Step~2 of Figure~\ref{fig:pipeline}, to transform generic Wikipedia paragraphs into task-specific samples, we use a prompt-based rewriting approach with the \texttt{ChatGPT-4o-mini}~\citep{achiam2023gpt}. All tasks follow the principle of information density to produce knowledge-rich outputs, with MCQ, TFQ, and CS pairs differing only in prompt formats (see Appendix~\ref{sec:prompt}). 
For paragraph-level tasks, segmented text blocks that are correctly parsed and meet template constraints are directly used as high-quality samples; invalid ones are discarded.

Following previous work~\citep{li2024llms,chen2025safeeraser,pang2024improving}, we further use \texttt{ChatGPT-4o-mini} to perform quality scoring along multiple dimensions, including rarity, complexity, and informativeness. The scores are aggregated into a single overall rating, which is then discretized to a 1-5 scale. 
To ensure consistency across tasks and sessions, we retain only samples with a score of 4 or above as the final high-quality subset, while those below 4 are regarded as low-quality data.

\textbf{Low-Quality Sample Generation.}
To enrich alignment signals, we use \texttt{ChatGPT-4o-mini} to generate multiple degraded variants for each high-quality sample, reducing information density or reasoning completeness while preserving the topic. 
Instead of using real low-quality data, which often lacks paired high-quality supervision, we construct synthetic variants to enable simple and controllable aligned training. 
For each target \(Y\), we generate \(k \in \{2,\dots,20\}\) variants (Figure~\ref{fig:split_count}), forming hierarchical mappings that encourage information aggregation and reasoning completion. We also include chain-of-thought as intermediate supervision. 
Appendix~\ref{sec:append_compare_rl} show that the synthetic data reasonably approximates real low-quality distributions.

\textbf{Robustness Enhancement.} To increase diversity and robustness, we perform cross-topic synthesis by selecting semantically similar sample pairs with constrained entity overlap, followed by GPT-based topic-fusion rewriting, retaining only samples with quality scores above 4. To further improve generalization to noisy inputs, we inject surface perturbations such as spelling variations, synonym substitutions, and minor formatting shifts during training, while keeping some clean samples to balance robustness and fidelity.

\section{LM-Mixup}
\label{sec:method}

This section is organized into three parts: (\romannumeral1) Cold Start Pretraining, (\romannumeral2) Multi-Dimensional Reward Design, and (\romannumeral3) Reinforcement Learning with GRPO. The overall framework of the proposed \method training pipeline, built upon the Qwen-2.5-1.5B-Instruct~\citep{team2024qwen2}, is illustrated in Figure~\ref{fig:training_pipeline}.

\subsection{Cold Start}
As directly applying reinforcement learning from random initialization is often unstable \citep{guo2025deepseek,wei2025advancing}, we first perform a cold-start supervised training stage on a subset of \dataset. 
For each example, the $k$ low-quality inputs are linearized as the conditional context, and the model is trained to generate the corresponding high-quality output with the standard autoregressive cross-entropy loss. 
This stage provides a stable initial policy for subsequent reinforcement learning.

\subsection{Reward Design}
\label{sec:reward_design}
To encourage the model to produce outputs with stronger information aggregation, semantic alignment, and structural conformity when mapping multiple low-quality samples to high-quality outputs, we design three complementary reward components. Given a model output $\hat{Y}$ and the corresponding high-quality reference $Y$, the total reward is defined as
\begin{equation}
    R(\hat{Y}, Y) = 
    \lambda_q R_q(\hat{Y}) + 
    \lambda_a R_a(\hat{Y}, Y) + 
    \lambda_f R_f(\hat{Y}),
    \label{eq:total_reward}
\end{equation}
where $\lambda_q, \lambda_a, \lambda_f$ are the normalized weights. In our experiments, we set $\lambda_q = 0.5$, $\lambda_a = 0.4$, and $\lambda_f = 0.1$.

\textbf{(1) Quality Reward $R_q$:} 
To efficiently approximate LLM ratings during training, we introduce a KNN-Bayes scoring scheme.  
Given a generated output $\hat{Y}$, we retrieve its $k$ nearest neighbors from a large reference set with pre-computed LLM scores and estimate the posterior distribution of the true quality label via a score transition matrix $T$~\citep{zhu2021clusterability,pang2024improving}:
\begin{equation}
    P(y=i \mid \mathbf{h}(\hat{Y})) 
    \propto p_i \cdot \exp\!\left(\sum_{j} h_j(\hat{Y}) \log T_{ij}\right),
    \label{eq:bayes_posterior_short}
\end{equation}
where $\mathbf{h}(\hat{Y})$ is the neighbor rating histogram. 
The expected quality score $\hat{s}(\hat{Y})$ from this posterior is then mapped into a parameterized piecewise reward:
\begin{equation}
R_q\big(\hat{Y};\,\lambda,\kappa,\alpha,\beta\big) = 
\begin{cases}
\alpha & \hat{s}(\hat{Y}) \ge \lambda,\\[2pt]
\beta & \hat{s}(\hat{Y}) = \kappa,\\[2pt]
0 & \text{otherwise},
\end{cases}
\label{eq:quality_reward_short}
\end{equation}

By default we set $\lambda{=}4$, $\kappa{=}3$, and $\alpha{=}1$, $\beta{=}0.3$. 
The offline construction of the reference set and transition matrix estimation is provided in Appendix~\ref{sec:detail_knn}.

\textbf{(2) Semantic Alignment Reward $R_a$:}
To ensure semantic consistency between generated outputs and reference answers, we encode both using the embedding model\footnote{\url{https://huggingface.co/BAAI/bge-m3}} and compute the normalized cosine similarity
\begin{equation}
    R_a(\hat{Y}, Y) = \mathbb{1}\left(\mathrm{cosine}(e(\hat{Y}), e(Y)) \ge \tau\right),
\end{equation}
where $e(\cdot)$ denotes the SentenceBERT encoder, $\tau$ is the similarity threshold, and $\mathbb{1}(\cdot)$ is the indicator function that returns $1$ if the condition holds and $0$ otherwise.

\textbf{(3) Format Compliance Reward $R_f$:}
To enforce structural consistency with the: \texttt{<think>...</think><answer>...</answer>} template, we use regular expressions to verify the output format. 
Outputs fully matching the template receive $R_f(\hat{Y})=1$, otherwise $0$.

Finally, the total reward in Eq.~\eqref{eq:total_reward} integrates quality, semantic alignment, and format compliance into a unified multi-dimensional signal.

\subsection{Reinforcement Learning with GRPO}

Building on cold-start pretraining and designed rewards, we adopt GRPO for reinforcement learning fine‐tuning. 
Unlike standard SFT, which forces the model to imitate a single reference answer, our \task allows infinitely many valid aggregation or generation strategies. 
Sole reliance on SFT risks overfitting to one canonical form and ignoring the diverse space of high‐quality outputs. 
In contrast, reinforcement learning enables optimizing directly against reward signals, encouraging exploration of diverse outputs and progressively improving generation quality.

Specifically, GRPO is a variant of PPO~\citep{schulman2017proximal} that removes the need for a learned value (critic) function by replacing the baseline with group‐wise statistics. 
For each input \(X\), the model samples multiple candidate outputs 
\(\{\hat{Y}_1, \ldots, \hat{Y}_m\}\), which are scored by the multi‐dimensional reward \(R(\hat{Y}_i, Y)\). 
To reduce variance and mitigate scale inconsistency across candidates, GRPO computes a normalized reward within each group:
\begin{equation}
\tilde{R}_i = \frac{R(\hat{Y}_i, Y) - \mu_X}{\sigma_X + \epsilon},
\end{equation}
where \(\mu_X\) and \(\sigma_X\) are the mean and standard deviation of \(\{R(\hat{Y}_j, Y)\}_{j=1}^m\), and \(\epsilon\) is a small constant to ensure numerical stability. The policy optimization objective becomes:

\begin{equation}
\scalebox{0.88}{$
\begin{aligned}
\mathcal{L}_{\mathrm{GRPO}}(\theta)
= {} & \mathbb{E}_{X}\Bigl[
\min\bigl(
r_i(\theta)\,\tilde{R}_i,\;
\mathrm{clip}(r_i(\theta), 1-\epsilon_{\mathrm{clip}}, 1+\epsilon_{\mathrm{clip}})\,\tilde{R}_i
\bigr)
\Bigr] - \beta\,\mathrm{KL}\!\bigl(
\pi_\theta(\cdot\mid X)\,\|\,\pi_{\theta_0}(\cdot\mid X)
\bigr).
\end{aligned}
$}
\end{equation}
where $r_i(\theta) = \frac{\pi_\theta(\hat{Y}_i \mid X)}{\pi_{\theta_0}(\hat{Y}_i \mid X)}$ is the importance ratio between the current policy \(\pi_\theta\) and reference (old) policy \(\pi_{\theta_0}\); \(\epsilon_{\mathrm{clip}}\) is the PPO clipping parameter; \(\beta\) controls the strength of KL regularization to ensure stability~\citep{christiano2017deep}.

\subsection{Capacity-Constrained Clustering}

After GRPO training, the model can distill multiple low-quality samples into high-quality ones. For downstream tasks, we introduce a Capacity-Constrained Clustering method to automatically collect low-quality inputs with flexible control over cluster number and size, which also mitigates the severe imbalance or over-fragmentation issues often observed in standard clustering methods. Given a text collection \(\mathcal{D}=\{x_i\}_{i=1}^N\), we encode each sample into \(\mathbf{h}_i \in \mathbb{R}^d\) using a pre-trained encoder. A target capacity vector \(\mathbf{c} = (c_1,\ldots,c_K)\) is drawn from a truncated normal distribution with \(c_k \in [c_{\min}, c_{\max}]\). We then perform two-stage clustering: (i) run MiniBatchKMeans to obtain \(k\) initial cluster centers \(\{\mathbf{c}_k\}\); (ii) iteratively assign samples to the most similar clusters under capacity constraints, with a few refinement steps to ensure semantic compactness and balanced partitioning.

\section{Experiments}
\label{sec:experiments}

\begin{table}[t]
\centering
\begin{minipage}{0.55\linewidth}
\centering
\resizebox{\linewidth}{!}{
\begin{tabular}{lccccc c}
\toprule
Model & cs & mcq & para & qa & tfq & Avg \\
\midrule
LLaMA-3.1-8B-Instruct & 3.61 & 2.57 & 3.57 & 3.71 & 2.10 & 3.27 \\
LLaMA-3.2-3B-Instruct & 3.58 & 2.66 & 3.52 & 3.78 & 2.49 & 3.21 \\
DeepSeek-R1-Distill-Qwen-7B & 3.61 & 2.46 & 3.40 & 3.23 & 2.41 & 3.02 \\
Qwen-2.5-7B-Instruct & 3.70 & 2.77 & 3.58 & 3.53 & 2.57 & 3.28 \\
Qwen-2.5-1.5B-Instruct & 3.39 & 2.44 & 3.34 & 3.33 & 1.34 & 2.86 \\
GPT-4o-mini   & 3.81 & 2.86 & \textbf{3.69} & 3.64 & 2.61 & 3.37 \\
Qwen-2.5-1.5B-SFT & 3.54 & 3.25 & 2.82 & 3.73 & 3.05 & 3.28 \\
Qwen-2.5-7B-SFT   & 3.53 & 3.31 & 3.41 & 3.78 & 3.10 & 3.46\\
\midrule
\method & \textbf{3.85} & \textbf{3.55} & 3.31 & \textbf{4.17} & \textbf{3.32} & \textbf{3.66} \\
\bottomrule
\end{tabular}}
\caption{Performance comparison across different models on five tasks. The best results per column are highlighted in \textbf{bold}.}
\label{tab:mixture_results}
\end{minipage}%
\hspace{0.05\linewidth}
\begin{minipage}{0.35\linewidth}
\centering
\resizebox{\linewidth}{!}{
\begin{tabular}{lc}
\toprule
\textbf{Datasets} & \textbf{Data size}  \\
\midrule
Flan V2 & 100K \\
Open-Assistant 1 & 33K \\
WizardLM & 100K \\
Dolly & 15K \\
Stanford Alpaca & 52K \\
\midrule
Overall & 300K \\
\bottomrule
\end{tabular}}
\caption{Data pool statistics.}
\label{tab:simplified_data_pool}
\end{minipage}
\end{table}


\subsection{\dataset Experimental Results}

\textbf{Experimental Setup.} 
To comprehensively evaluate the performance of \method on the \dataset dataset, we conducted standardized experiments on the test set using a variety of models. 
Specifically, the experiments involved the following models: ChatGPT-4o-mini~\citep{achiam2023gpt}, Qwen-2.5-1.5B-Instruct, Qwen-2.5-7B-Instruct~\citep{team2024qwen2}, LLaMA-3.1-8B-Instruct, LLaMA-3.2-3B-Instruct~\citep{llama3}, DeepSeek-R1-Distill-Qwen-7B~\citep{guo2025deepseek}, Qwen-2.5-1.5B-SFT, Qwen-2.5-7B-SFT (obtained via supervised fine-tuning on the full \dataset).
All models were evaluated on the same test set, constructed by holding out a non-overlapping 20\% split from \dataset, under identical prompt conditions to ensure fair comparison.
For automated evaluation, we employed ChatGPT-4o-mini as the rating model to assess the quality of the generated outputs.

\textbf{Results.} 
Table~\ref{tab:mixture_results} compares different models on the \dataset test set. 
\method consistently outperforms all baselines and achieves the best overall results. Compared with standard supervised fine-tuning, GRPO with multi-dimensional quality rewards enables the model to learn higher-quality generation patterns rather than merely mimicking the ground truth.

\subsection{OpenLLM Leaderboard Evaluation Results}
\label{sec:openllm}

\begin{table}[h]
    \centering
    \resizebox{\linewidth}{!}{
    \begin{tabular}{l|cccccc}
    \toprule
    \multirow{2}{*}{\textbf{Model}} & \multicolumn{1}{c}{\textbf{MMLU}} & \multicolumn{1}{c}{\textbf{TruthfulQA}} & \multicolumn{1}{c}{\textbf{GSM}} & \multicolumn{1}{c}{\textbf{BBH}} & \multicolumn{1}{c}{\textbf{TydiQA}} & \multicolumn{1}{c}{\textbf{Average}} \\
    & (\textbf{factuality}) & (\textbf{truthfulness}) & (\textbf{reasoning}) & (\textbf{reasoning})  & (\textbf{multilinguality})  & \\
    \midrule
    \textsc{Vanilla base model*}    & 64.1 & 33.5 & 56.5 & 55.4 & 23.3 & 46.6 \\
    \textsc{Completion Length*}     & 64.2 & 41.4 & 62.5 & 60.7 & 23.0 & 50.4 \\
    \textsc{Perplexity*}            & 63.1 & 40.4 & 55.5 & 60.2 & 62.1 & 56.3 \\
    \textsc{$k$-NN-10*}             & 62.4 & 44.3 & 57.0 & 59.1 & 63.8 & 57.3 \\
    \textsc{Random Selection*}      & 63.4 & 39.1 & 62.2 & 61.3 & 61.1 & 57.4 \\
    \textsc{LESS*}                  & 63.0 & 39.0 & 57.5 & 63.1 & 67.2 & 58.0 \\
    \textsc{Full data (300K)*}      & 63.5 & 42.0 & 61.0 & 59.1 & 62.8 & 57.7 \\
    \midrule
    \textsc{AlpaGasus*}             & 63.4 & 42.6 & 66.0 & 59.1 & 59.4 & 58.1 \\
    \textsc{Deita*}                 & 64.5 & 50.1 & 60.0 & 60.3 & 63.7 & 59.7 \\
    \textsc{DS2 w/o curation*}     & 63.3 & 51.5 & 62.0 &59.7 & 64.3 & 60.2 \\
    \textsc{DS2*}                  & 64.0 & 50.3 & 67.5 & 59.0 & 66.1 & \underline{61.4} \\
    \midrule
    
    Back-translation
& 62.0{\footnotesize$\pm$0.4} & 46.5{\footnotesize$\pm$2.9} & 61.2{\footnotesize$\pm$0.8} & 58.8{\footnotesize$\pm$2.2} & 60.2{\footnotesize$\pm$0.8} & 57.7{\footnotesize$\pm$0.1} \\
EDA
& 61.6{\footnotesize$\pm$0.9} & 43.7{\footnotesize$\pm$2.0} & 56.2{\footnotesize$\pm$1.0} & 59.7{\footnotesize$\pm$0.3} & 62.0{\footnotesize$\pm$1.6} & 56.6{\footnotesize$\pm$0.6} \\
Rephrasing
& 61.4{\footnotesize$\pm$0.7} & 36.0{\footnotesize$\pm$2.5} & 63.2{\footnotesize$\pm$1.0} & 59.6{\footnotesize$\pm$0.2} & 62.2{\footnotesize$\pm$0.8} & 56.5{\footnotesize$\pm$0.6} \\
\midrule
\textsc{Base 70\% + Ori 30\%}
& 62.2{\footnotesize$\pm$0.9}
& 40.7{\footnotesize$\pm$0.5}
& 54.3{\footnotesize$\pm$0.2}
& 55.1{\footnotesize$\pm$1.1}
& 23.2{\footnotesize$\pm$0.3}
& 47.1{\footnotesize$\pm$0.1} \\

\textsc{Base 50\% + Ori 50\%}
& 62.1{\footnotesize$\pm$0.1}
& 37.4{\footnotesize$\pm$0.7}
& 50.8{\footnotesize$\pm$0.6}
& 54.1{\footnotesize$\pm$0.7}
& 22.9{\footnotesize$\pm$0.4}
& 45.4{\footnotesize$\pm$0.2} \\
\textsc{Base 30\% + Ori 70\%}
& 61.3{\footnotesize$\pm$0.7}
& 38.2{\footnotesize$\pm$0.7}
& 51.2{\footnotesize$\pm$1.4}
& 54.2{\footnotesize$\pm$0.9}
& 23.0{\footnotesize$\pm$0.5}
& 45.6{\footnotesize$\pm$0.4} \\

    \midrule
    \textsc{Low 70\% + Ori 30\%} & 62.7{\footnotesize$\pm$0.7}
& 17.8{\footnotesize$\pm$2.0}
& 62.5{\footnotesize$\pm$4.0}
& 60.3{\footnotesize$\pm$0.9}
& 65.6{\footnotesize$\pm$1.1}
& 53.6{\footnotesize$\pm$1.2} \\
    \rowcolor{gray!20}\textbf{\textsc{Mixup 70\% + Ori 30\%}} & 63.0{\footnotesize$\pm$0.2}
&47.9{\footnotesize$\pm$0.3}
& 63.3{\footnotesize$\pm$0.6}
& 61.1{\footnotesize$\pm$0.3}
& 64.2{\footnotesize$\pm$0.6}
& \textbf{59.9}{\footnotesize$\pm$0.1}\scriptsize\textcolor{customgreen}{↑6.3} \\
    \textsc{Low 50\% + Ori 50\%}   & 62.4{\footnotesize$\pm$0.6}
& 39.0{\footnotesize$\pm$9.8}
& 62.7{\footnotesize$\pm$1.1}
& 61.0{\footnotesize$\pm$2.2}
& 64.0{\footnotesize$\pm$0.3}
& 57.9{\footnotesize$\pm$2.0} \\
    \rowcolor{gray!20}\textbf{\textsc{Mixup 50\% + Ori 50\%}} & 63.3{\footnotesize$\pm$0.3}
& 52.6{\footnotesize$\pm$0.1}
& 65.5{\footnotesize$\pm$0.2}
& 61.3{\footnotesize$\pm$0.3}
& 64.6{\footnotesize$\pm$0.3}
& \textbf{61.5}{\footnotesize$\pm$0.1}\scriptsize\textcolor{customgreen}{↑3.6} \\
    \textsc{Low 30\% + Ori 70\%} & 60.9{\footnotesize$\pm$2.1}
& 41.1{\footnotesize$\pm$5.6}
& 62.7{\footnotesize$\pm$1.9}
& 59.9{\footnotesize$\pm$1.7}
& 60.4{\footnotesize$\pm$1.7}
& 57.0{\footnotesize$\pm$0.8} \\
    \rowcolor{gray!20}\textbf{\textsc{Mixup 30\% + Ori 70\%}} & \textbf{63.1}{\footnotesize$\pm$0.3}
& 46.8{\footnotesize$\pm$0.6}
& 61.2{\footnotesize$\pm$1.5}
& 58.0{\footnotesize$\pm$0.2}
& 63.4{\footnotesize$\pm$1.1}
& \textbf{58.5}{\footnotesize$\pm$0.1} \scriptsize\textcolor{customgreen}{↑1.5}\\
    \bottomrule
    \end{tabular}
    }
        \caption{Results on the OpenLLM leaderboard using \texttt{LLaMA-3.1-8B} as the base model. The top-performing scores are shown in \textbf{bold}, while the second-best scores are marked with \underline{underlines}. Unless otherwise specified, the size of the fine-tuning dataset is 10K. * indicates that the values are sourced from \citet{pang2024improving}.}
    \label{tab:openllm_results_base_llama3.1}
\end{table}

\begin{figure}[t]
    \centering
    \begin{subfigure}[t]{0.48\textwidth}
        \centering
        \includegraphics[width=0.75\linewidth]{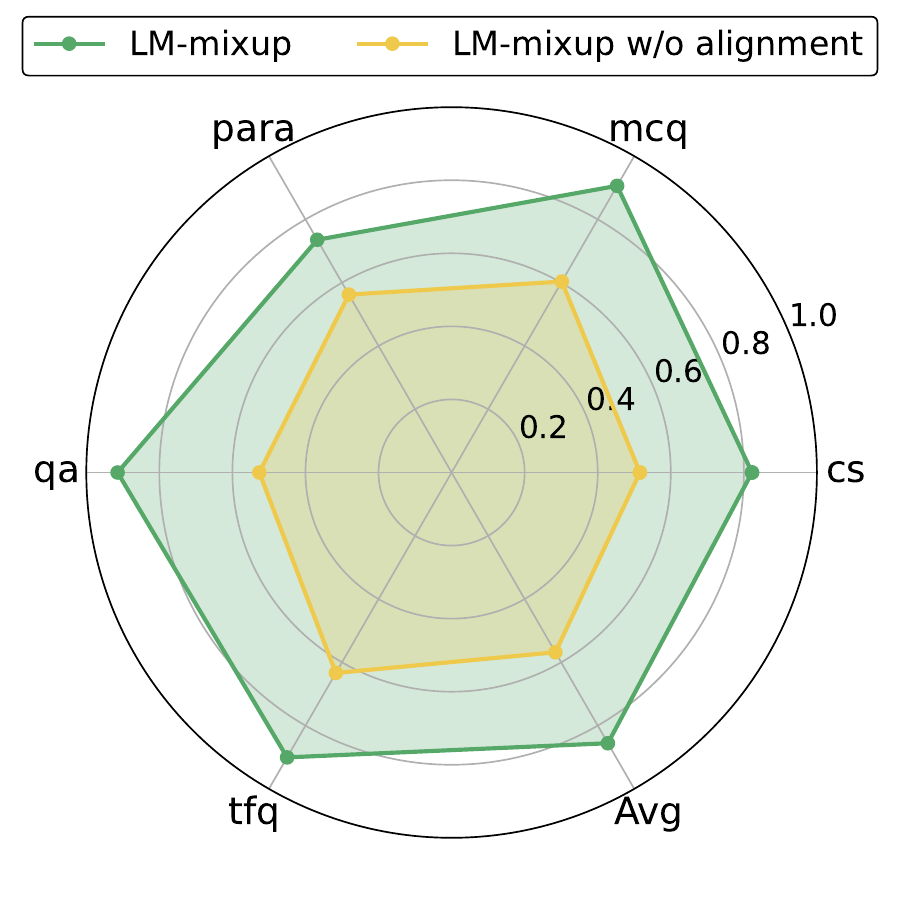}
    \end{subfigure}
    \hfill
    \begin{subfigure}[t]{0.48\textwidth}
        \centering
        \includegraphics[width=\linewidth]{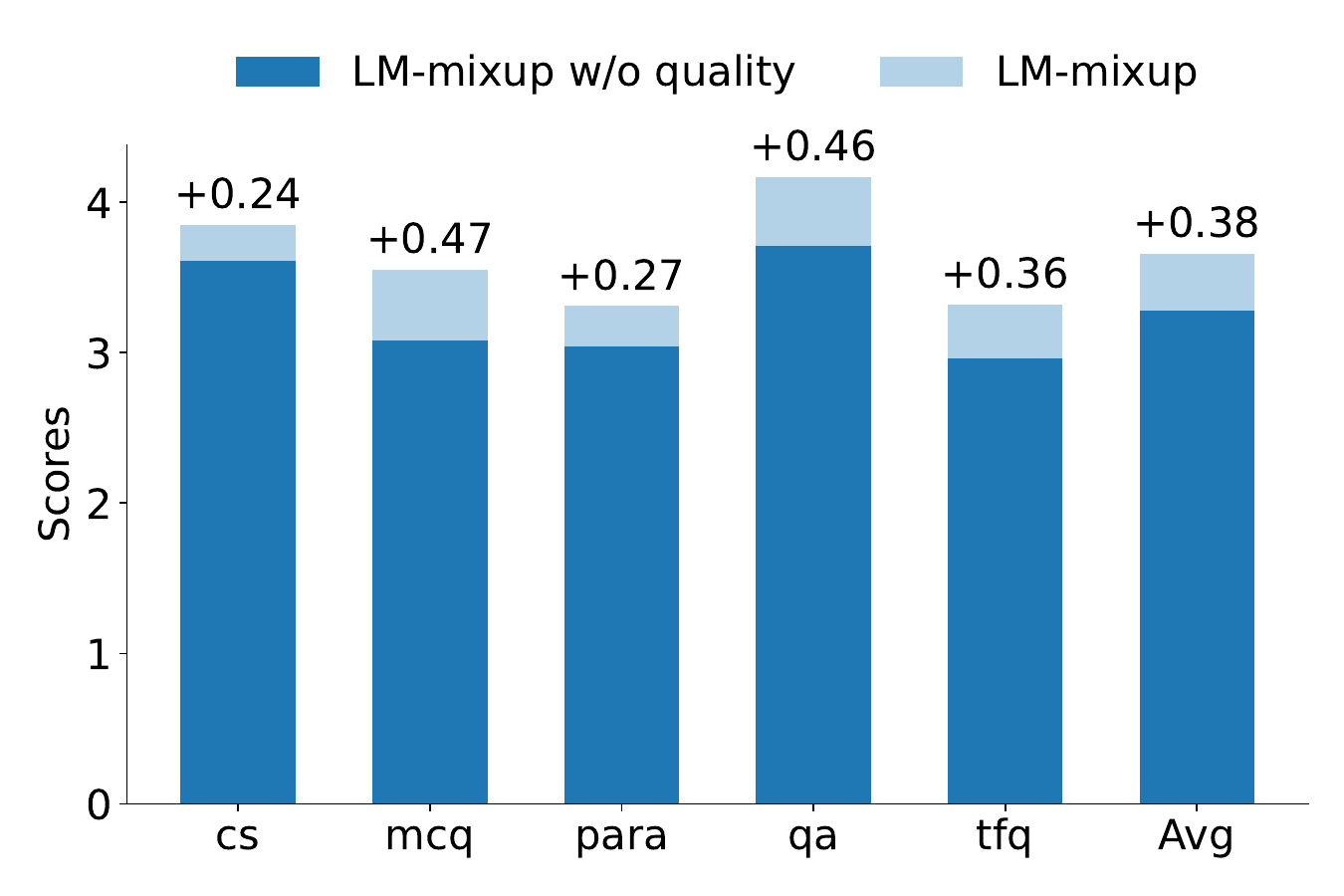}
    \end{subfigure}
    \vspace{-3mm}
    \caption{Ablation study on reward components in \method. 
    The left figure evaluates the effect of removing the alignment reward, 
    while the right figure shows the impact of removing the quality reward across different tasks.}
    \label{fig:ablation_rewards}
\end{figure}

\begin{figure}
    \centering
    \begin{minipage}{0.5\linewidth}
        \centering
        \includegraphics[width=\linewidth]{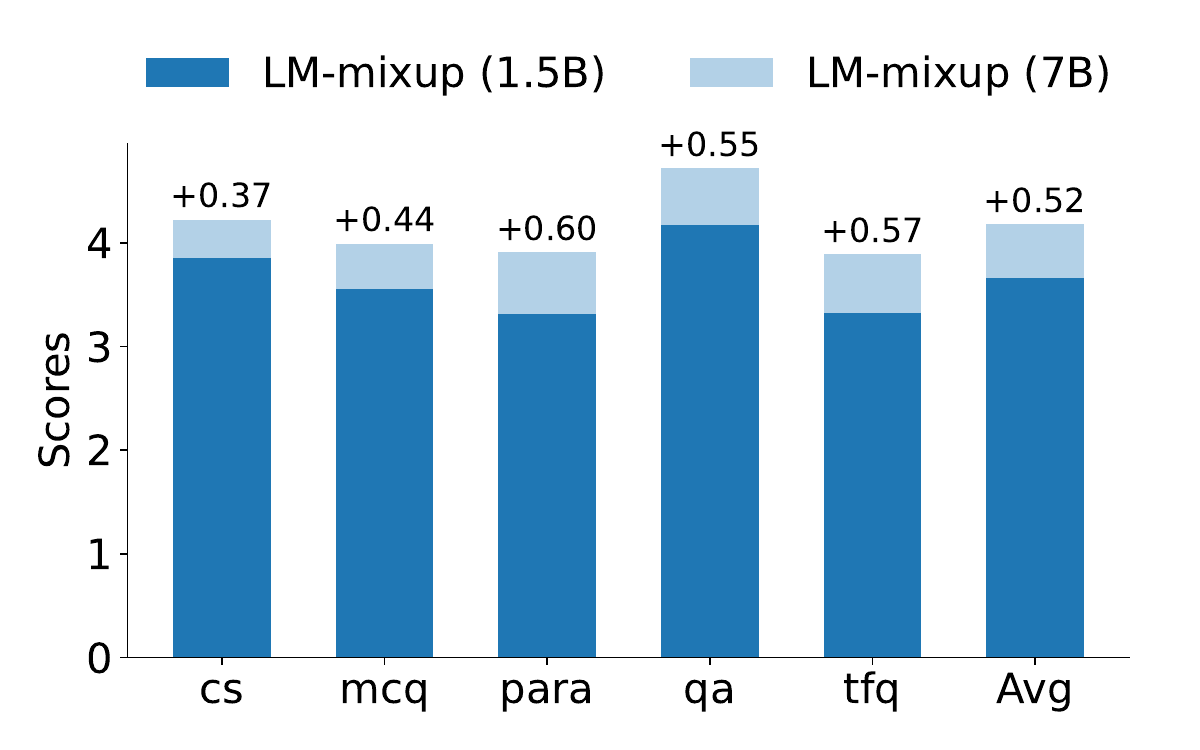}
        \caption{Effect of model scaling on performance.}
        \label{fig:scaling}
    \end{minipage}
    \hfill
    \begin{minipage}{0.48\linewidth}
        \centering
        \includegraphics[width=\linewidth]{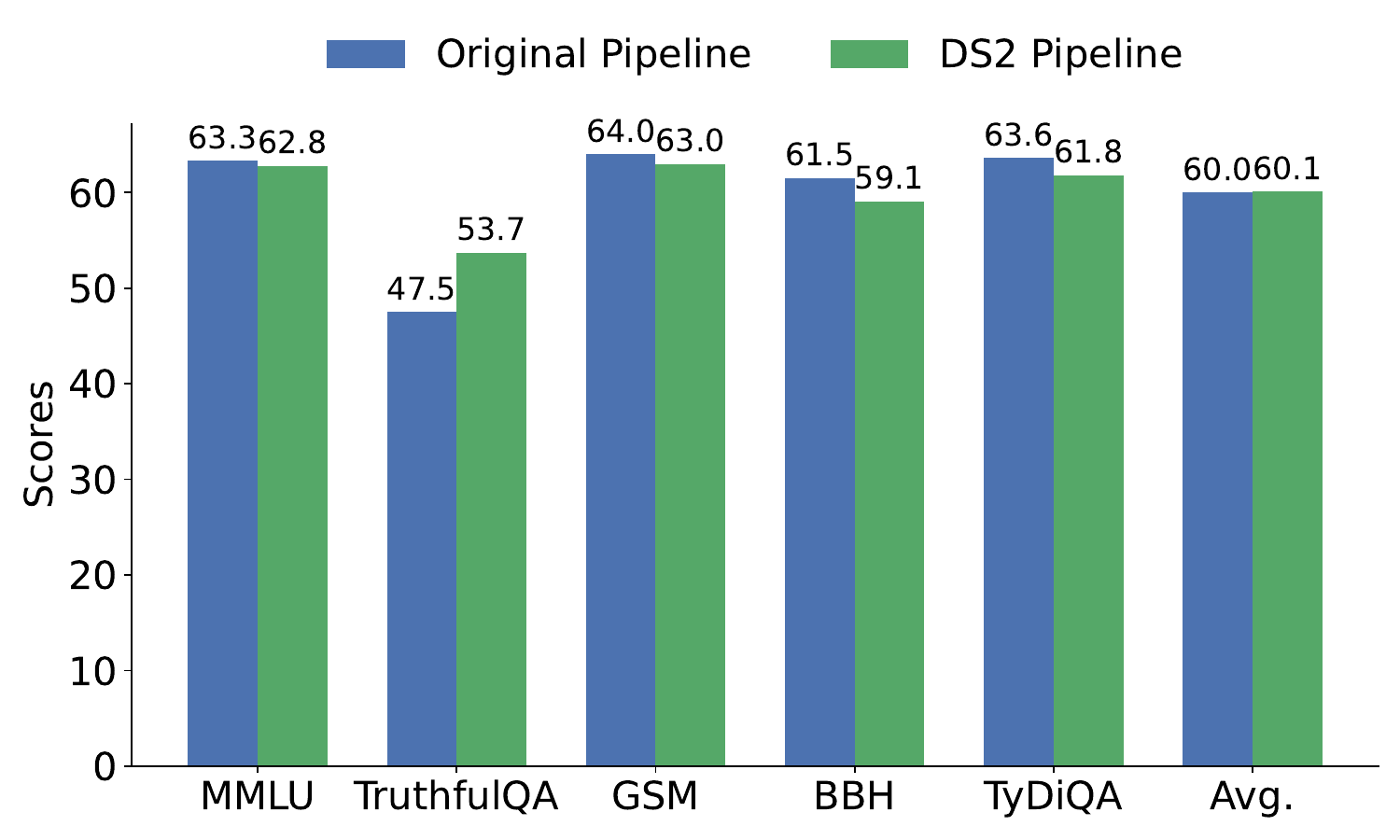}
        \caption{Comparison of the \method using DS2 pipelines under the Mixup 70\% + Ori 30\%.}
        \label{fig:ds2}
    \end{minipage}
\end{figure}

\textbf{Experimental Setup.} 
Following previous work~\citep{pang2024improving}, to evaluate \method’s performance on OOD datasets, we construct an additional data pool consisting of Flan\_v2~\citep{longpre2023flan}, Open Assistant 1~\citep{kopf2023openassistant}, WizardLM~\citep{xu2023wizardlm}, Dolly~\citep{dolly}, and Stanford Alpaca~\citep{stanford}. 
Detailed statistics of the data pool are provided in Table~\ref{tab:simplified_data_pool}. To identify low-quality samples within this pool, we employ \texttt{ChatGPT-4o-mini} for quality rating following the same protocol described in Sec~\ref{sec:high_qual_gen}, where samples with a score below 4 are regarded as low-quality data.
We then apply \method to perform mixup on the low-quality samples within the data pool and compute long-tail scores using embeddings. 
The top-ranked samples from both the original high-quality data and the mixup data from low-quality data are then selected for instruction fine-tuning.

\textbf{Metrics.} We report task-specific metrics, including accuracy on MMLU~\citep{mmlu}, BBH~\citep{bbh}, and GSM8K~\citep{gsm}, the Informative-Truthful Rate on TruthfulQA~\citep{lin2021truthfulqa}, and F1 scores on TyDiQA~\citep{clark2020tydi}.

\textbf{Training Settings.} 
We fine-tune three base models, LLaMA-2-7B~\citep{llama2}, LLaMA-3.1-8B~\citep{llama3}, and Mistral-7B-v0.3~\citep{jiang2023mistral7b}, on 10K samples under multiple data settings. 
Specifically, we consider \textbf{ORI} (high-quality data only), \textbf{LOW} (low-quality data only), \textbf{BASE} (zero-shot generation with Qwen-1.5B-Instruct), and \textbf{MIXUP} (data generated by our LM-Mixup pipeline). Further details are provided in Appendix~\ref{sec:other_setting}.
For mixed-data training, we combine ORI with LOW, BASE, or MIXUP while keeping the total training-set size fixed at 10K. The relative proportion between the original high-quality data and the additional data source can be regarded as a data-composition hyperparameter, which may be adjusted according to the quality, diversity, and domain characteristics of the available data. Rather than exhaustively tuning this ratio for each model or benchmark, we evaluate three representative settings, with the additional data accounting for 30\%, 50\%, or 70\% of the training set. These settings respectively represent original-data-dominant, balanced, and additional-data-dominant regimes. We report the mean over three runs for our experiments and use published results for prior baselines.
We additionally report results on the full MIXUP and full LOW settings in Appendix~\ref{sec:appen_exper}.

\textbf{Baselines.} 
We compare our method against several representative data selection baselines commonly used in LLM fine-tuning, including \textit{Random Selection}, \textit{Completion Length}, \textit{Perplexity}, \textit{k-NN}, \textit{LESS}~\citep{xia2024less}, \textit{AlpaGasus}~\citep{chen2023alpagasus}, \textit{DEITA}~\citep{deita}, \textit{DS2}~\citep{pang2024improving}, and \textit{Full Data}.  We also include widely used data augmentation baselines, including back-translation~\citep{backtrans}, paraphrasing~\citep{rephrase}, and EDA~\citep{wei2019eda}. Additionally, to more comprehensively assess the gains achieved by \method, we also report the zero-shot performance of Qwen-1.5B-Instruct without any further training.
Detailed descriptions of these baselines are provided in Appendix~\ref{app:baseline_details}.

\textbf{Low-quality data enhanced by \method can even outperform high-quality-only baselines.}
Table~\ref{tab:openllm_results_base_llama3.1} shows that combining low-quality data (after mixup) with original high-quality samples can surpass baselines that rely solely on high-quality data selection. In particular, the 50\% MIXUP + 50\% ORI configuration achieves the top average score across all five OpenLLM Leaderboard benchmarks, with Tables~\ref{tab:openllm_results_base_mistral} and \ref{tab:openllm_results_base_llama2} showing similar trends on Mistral-7B and LLaMA-2-7B. This demonstrates that even low-quality data, when fused into high-quality samples, can enhance diversity and complement real data to boost performance. Additional results on more models are provided in the Appendix \ref{sec:appen_exper}. 

\textbf{\method demonstrates strong performance.} As shown in Table~\ref{tab:openllm_results_base_llama3.1}, \method significantly outperforms the zero-shot Qwen-1.5B-Instruct model used for mixup without any training. 
Furthermore, our method consistently surpasses standard data augmentation baselines, indicating that it truly elevates low-quality samples into high-quality supervision signals rather than merely increasing surface-level diversity.

\textbf{Mixup outperforms the full data pool with only 3.3\% of the data.}
For both LLaMA-3.1-8B (Table~\ref{tab:openllm_results_base_llama3.1}) and Mistral-7B (Table~\ref{tab:openllm_results_base_mistral}), our best-performing mixup configuration using only 10K training samples even surpasses the 300K full data pool baseline, demonstrating that mixup not only enhances data diversity but also enables a highly compact training set to outperform large-scale unfiltered data.

\textbf{Effect of mixup on low quality data.} 
As shown in Table~\ref{tab:openllm_results_base_llama3.1}, applying \method to low-quality data consistently improves performance across all mixture ratios.  
E.g., in the 70\% LOW + 30\% ORI setting, \method raises the score from 53.6 to 59.9 ($\uparrow$6.3), with similar gains in the 50\% ($\uparrow$3.6) and 30\% ($\uparrow$1.5) settings.  
This highlights that properly modeling low-quality data can yield substantial benefits for model training.

\subsection{Ablation Study}
\textbf{Ablation on Reward Components.} We conduct ablation studies to investigate the contribution of each reward component in our GRPO-based \method training framework, which incorporates quality and alignment rewards alongside the base objective. As shown in Figure \ref{fig:ablation_rewards}, removing the alignment reward causes the model to exhibit reward hacking behavior: it tends to memorize answers from the reference set regardless of the input, leading to significantly lower semantic similarity with the ground truth. On the other hand, removing the quality reward makes the model behave similarly to standard SFT, producing outputs with limited quality improvement. These highlight that both rewards are essential: the alignment reward ensures semantic faithfulness to the input, while the quality reward drives the generation of high-quality outputs beyond simple imitation.

\textbf{Effect of Model Scaling.} To investigate the impact of scaling up model parameters, we extend our training pipeline from Qwen-2.5-1.5B-Instruct to Qwen-2.5-7B-Instruct using the same GRPO-based optimization described in Sec.\ref{sec:method}. As shown in Fig.\ref{fig:scaling}, the larger 7B model consistently outperforms its 1.5B counterpart across all tasks on the \dataset test set, achieving an average score of 4.18 compared to 3.66 on the smaller model. These results demonstrate both the effectiveness and the scalability of our approach when applied to models with larger parameter sizes.

\textbf{Revisiting LLM rating bias.}
Recent work has noted that LLM-as-judge scores can be biased~\citep{ye2024justice,chen2024humans}. 
In our pipeline we use \texttt{ChatGPT-4o-mini} for rating, which may introduce such bias. 
To assess sensitivity, we conducted experiments using the DS2 pipeline~\citep{pang2024improving}. 
We conduct the same experiments described in Sec.\ref{sec:openllm} experiments under the MIXUP 70\% + ORI 30\% setting, where the overall performance shows only marginal changes, as shown in Fig.\ref{fig:ds2}. We hypothesize two reasons: (i) \method’s GRPO with multi-dimensional rewards and many-to-one mixup supervision provides strong signals that attenuate upstream rating noise; and (ii) diversity is governed by embedding-based long-tail selection, largely independent of the rating scale. Overall, while LLM rating bias is real, our design appears tolerant to moderate bias; further de-biasing (e.g., multi-judge ensembling, cross-model adjudication, or light human spot-checks) may be needed to unlock additional gains.

\section{Conclusion}
In this work, we introduce \task, a new setting for converting multiple low-quality, noisy, or redundant inputs into a single high-quality output, together with \dataset, a large-scale dataset for this task. We further propose \method, which combines supervised initialization and GRPO-based optimization with customized rewards. Results show that \method can turn imperfect data into effective supervision, improving both data efficiency and the performance of instruction-tuned LLMs.

\section*{Acknowledgments}
ZD, LL, JW are partially supported by the Yangcheng Scholars Research Project (No.~2024312049), CNPC Technology Project "Research on Key Technologies of Artificial Intelligence for Oil and Gas Exploration and Development" (2023DJ84) and the Guangdong Provincial Key Laboratory of Integrated Communication, Sensing and Computation for Ubiquitous Internet of Things (No.~2023B1212010007).

\bibliography{colm2026_conference}
\bibliographystyle{colm2026_conference}

\clearpage
\appendix
\startcontents[appendix]

\begin{center}
    {\LARGE\bfseries Appendix}
\end{center}

\vspace{3mm}

\section*{Table of Contents}
\hrule
\printcontents[appendix]{}{1}{
    \setcounter{tocdepth}{2}
}

\vspace{5mm}
\hrule
\vspace{8mm}

\section{Conceptual Distinction from Data Augmentation and Data Curation}
\label{subsec:distinction}

A central goal of our work is to address the pervasive scarcity of high-quality supervision in many NLP scenarios, especially for low-resource or domain-specific tasks. In such settings, large quantities of noisy, redundant, or otherwise low-quality instruction–response pairs are often available, whereas carefully curated high-quality data are expensive and limited. Our instruction distillation framework aims to transform abundant low-quality signals into a compact set of high-quality supervision examples, thereby improving the utility of existing corpora for downstream instruction tuning and supporting the broader low-resource NLP community.

Concretely, the MIXTURE dataset defines five heterogeneous task types. As shown in Sec.~5.1, LM-mixup achieves consistently strong performance across these tasks, indicating that the proposed instruction distillation paradigm is not tied to a single setting or a narrow engineering trick, but rather exhibits robustness and generality across diverse instruction-following scenarios.

Conceptually, our notion of \task is distinct from both traditional data augmentation and data curation:

\begin{itemize}
    \item \textbf{Instruction distillation.}
    Given multiple low-quality or inconsistent responses, instruction distillation extracts the useful information across them and semantically fuses these weak signals into a single, high-quality instruction–response pair. This process simultaneously aggregates information, denoises spurious content, and enforces quality-control constraints, yielding supervision with higher information density for downstream models.

    \item \textbf{Data augmentation.}
    Classical augmentation techniques generate additional samples via transformations such as rewriting~\citep{wei2019eda}, paraphrasing~\citep{rephrase}, back-translation~\citep{backtrans}, or other synthetic procedures, with the primary goal of expanding data volume and increasing diversity and robustness~\citep{aug1, aug2}. The underlying semantic content of each example is usually preserved, and the number of samples grows.

    \item \textbf{Data curation.}
    Data curation typically focuses on improving annotation quality or consistency for existing samples without substantially changing their semantic content~\citep{cur1,cur2}. Examples include relabeling noisy instances, filtering problematic examples, or correcting minor errors while keeping the original instruction–response structure intact.
\end{itemize}

Instruction distillation fundamentally differs from these two paradigms in both direction and effect. Instead of increasing the number of samples, it \emph{reduces} data volume via information aggregation, while \emph{changing and enriching} the semantic content through fusion across multiple weak sources. Unlike augmentation, which primarily improves diversity, or curation, which mainly refines labels for fixed content, instruction distillation explicitly converts many low-quality signals into a few high-information-density instructions. This enables substantial gains in low-resource regimes, where the key bottleneck is not the absolute number of examples, but the lack of sufficiently rich and reliable supervision.

\section{Dataset Construction and Analysis}
\subsection{Additional Details of the Dataset Construction Pipeline}
\label{sec:add_details_mixture}
\textbf{Data Cleaning and Preprocessing.}
In the initial stage, we extract structured entries from Wikipedia and perform systematic cleaning and segmentation to ensure the stability and controllability of subsequent task construction. Specifically, we retain only plain text content by removing templates, citation markers, tables, and other special formatting elements, and filter out excessively short entries. In practice, we set a minimum article length of 200 tokens to avoid information sparsity, particularly for paragraph-based tasks.

To reduce redundancy, we adopt a two-stage deduplication strategy. We first apply exact string matching to remove identical entries, and then perform near-duplicate filtering using Sentence-BERT, discarding samples with cosine similarity above 0.95. This process helps improve data diversity while preserving semantic coverage.

\textbf{Paragraph Segmentation.}
For segmentation, we first split each article into sentences based on punctuation such as periods, question marks, and exclamation marks. We then employ a greedy strategy to concatenate consecutive sentences into length-controlled segments. Specifically, we set a maximum length of \(T = 512\) tokens and allow approximately 10\% overlap between adjacent segments to preserve local context continuity. Segments shorter than 128 tokens are merged with neighboring segments to avoid introducing excessively short samples, while overly long sentences are further split using weaker punctuation (e.g., commas and semicolons). 

Through this process, we obtain a set of text segments that are balanced in length, semantic coherence, and distribution, providing a reliable foundation for downstream multi-task data construction.

\textbf{Task Construction.}
Based on the processed corpus, we construct five types of tasks to simulate diverse information organization patterns and provide supervision for aggregating multiple low-quality inputs into a single high-quality output. The task types include QA, MCQ, TFQ, CS, and Paragraph.

Despite differences in output formats, all tasks share a common objective: integrating multiple fragmented, low-quality, incomplete, or poorly structured inputs into a coherent and high-quality target. Specifically, the QA task aggregates multiple low-quality question–answer pairs into a refined QA pair; MCQ combines scattered multiple-choice-related fragments into a well-formed multiple-choice question; TFQ synthesizes low-quality true/false statements into a consistent judgment task; CS constructs structured category–statement pairs; and Paragraph generates a coherent and information-rich paragraph from multiple low-quality text segments. 

In essence, these tasks differ mainly in their output formats, while all modeling the same underlying process of multi-source information aggregation.
\subsection{Comparison with Real Low-Quality Data}
\label{sec:append_compare_rl}
The proposed \task aims to synthesize high-quality outputs from multiple low-quality inputs. Instead of directly collecting real low-quality data, we adopt a more controllable construction strategy by generating low-quality inputs from existing high-quality samples using \texttt{GPT-4o-mini}. The key motivation is that, although real low-quality data is abundant in open corpora, it typically lacks aligned high-quality supervision under the same semantic target, making it difficult to form stable training pairs for supervised learning. In contrast, generating semantically related but degraded inputs from high-quality samples ensures a clear correspondence between inputs and targets, while significantly reducing data construction cost. This makes it more suitable for studying low-to-high quality information aggregation.

To assess how well the synthetic low-quality data reflects the statistical properties of real low-quality text, we conduct a distributional alignment analysis, systematically comparing synthetic and real data across several key dimensions.

\textbf{Experimental Setup.}
We select two real-world low-quality corpora as references: Yahoo Answers\footnote{\url{https://ciir.cs.umass.edu/downloads/nfL6/nfL6.json.gz}} and Reddit\footnote{\url{https://huggingface.co/datasets/webis/tldr-17}}. Since different real corpora naturally exhibit distributional variations, we further include a real–real comparison (Yahoo Answers vs. Reddit) as a reference baseline. This allows us to examine whether the synthetic–real gap falls within the natural variation observed across real low-quality data sources.

We evaluate three task-relevant metrics: \emph{semantic cosine similarity}, \emph{type-token ratio (TTR)}~\citep{richards1987type},  and \emph{bigram entropy}~\citep{shannon1948mathematical}. Semantic cosine similarity measures redundancy across low-quality fragments; TTR captures lexical diversity; and bigram entropy reflects local information richness and compositional variability. We compute each metric independently for each dataset and perform pairwise comparisons, interpreting the synthetic--real gap relative to the natural variation observed between real-world corpora.

\begin{table}[t]
\centering
\resizebox{0.7\textwidth}{!}{
\begin{tabular}{lccc}
\toprule
\textbf{Metric} & \textbf{Yahoo--Reddit} & \textbf{Syn--Yahoo} & \textbf{Syn--Reddit} \\
\midrule
Semantic CosSim & 18.1\% & 14.1\% & 4.5\% \\
TTR             & 1.5\%  & 1.6\%  & 0.8\% \\
Bigram Entropy  & 4.1\%  & 6.9\%  & 2.3\% \\
\bottomrule
\end{tabular}
}
\caption{Distributional gap comparison between synthetic and real low-quality data. Lower values indicate smaller divergence.}
\label{tab:distribution_gap}
\end{table}

\textbf{Results.}
Table~\ref{tab:distribution_gap} show that the synthetic low-quality data exhibits strong alignment with real low-quality text across multiple dimensions. In particular, the gaps in TTR and bigram entropy are consistently small, indicating comparable lexical diversity and information density. For semantic cosine similarity, the differences between synthetic and real data also remain limited, suggesting that the redundancy structure among fragments is well captured. More importantly, when comparing the synthetic–real gap with the real–real gap, we find that the former largely falls within the natural variation observed between different real corpora. This suggests that the proposed synthetic data effectively captures key statistical characteristics of real low-quality inputs.

\section{Details of KNN--Bayes Rating}
\label{sec:detail_knn}
\subsection{KNN--Bayes Quality Modeling with Score Transition Matrix}

In the Sec.\ref{sec:reward_design}, we introduced an offline KNN--Bayes calibration method to approximate the original LLM ratings during training. Intuitively, given the $k$-nearest neighbors of each sample in the embedding space, we aim to infer its ``true'' quality score based on the observed ratings of these neighbors. However, the LLM-provided scores $\tilde y$ typically suffer from systematic noise and random fluctuations. Directly averaging the neighbor scores may therefore introduce significant bias into the reward signal.

To address this issue, we adopt the classical idea of Score Transition Matrix (STM) from weak supervision and noisy-label learning, which models the conditional distribution between observed and latent labels. Let the latent true label be $y \in \mathcal{Y} = \{1,2,\dots,C\}$ and the observed noisy rating be $\tilde y$. In our implementation, we set $C=6$ with label set $\{0,1,2,3,4,5\}$, which matches the original data annotation. The STM is defined as
\begin{equation}
    T \in \mathbb{R}^{C\times C}, 
    \qquad T_{ij} = \mathbb{P}(\tilde y = j \mid y = i),
\end{equation}
where $T_{ij}$ denotes the probability that a true label $i$ is perturbed into the noisy label $j$. The prior distribution is given by
\begin{equation}
    p \in \Delta^C, 
    \qquad p_i = \mathbb{P}(y=i), 
    \quad \sum_i p_i = 1.
\end{equation}
When $T = I$, the observed ratings are noise-free; deviations of $T$ from the identity matrix characterize systematic label noise.

\paragraph{$k$-NN Clusterability Assumption~\citep{wei2020combating}.}  
In the embedding space, if $x'$ belongs to the $k$-nearest neighbors $\mathcal{N}_k(x)$ of $x$, then it is more likely that $y(x') = y(x)$. Based on this assumption, the neighborhood agreement frequencies yield a set of linear equations over $(T,p)$. We adopt 2-NN consensus statistics when estimating $(T,p)$ to ensure identifiability. For the posterior computation of a single sample, we use the $k$-nearest neighbor histogram $h(x)$ with $k \ge 2$ to enhance robustness. Specifically, using pairwise or triplet neighbor agreement, we define
\begin{equation}
    v^{[1]} = T^\top p,
    \qquad v^{[2]}_\ell = (T \circ T_\ell)^\top p,
    \qquad v^{[3]}_{\ell,s} = (T \circ T_\ell \circ T_s)^\top p,
\end{equation}
where $T_\ell = T A_\ell$ is the cyclic shift of $T$ by $\ell$ units, and $\circ$ denotes the Hadamard product. The observed frequencies $\widehat{v}^{[1]}, \widehat{v}^{[2]}_\ell, \widehat{v}^{[3]}_{\ell,s}$ can be directly computed from data, forming a linear program over $(T,p)$.  We solve for $(T,p)$ subject to $T\mathbf{1}=\mathbf{1}$, $T\ge 0$, $p\ge 0$, and $\mathbf{1}^\top p=1$.
Existing theory shows that under mild identifiability conditions, third-order consensus vectors suffice to uniquely recover $(T,p)$.

Once $(T,p)$ are estimated, given the empirical neighbor histogram $h_j(x)$ of sample $x$, the posterior distribution is computed as
\begin{equation}
    \mathbb{P}(y=i \mid h(x)) 
\propto p_i \prod_{j\in\mathcal{Y}} T_{ij}^{h_j(x)}
= p_i \exp\!\left(\sum_{j\in\mathcal{Y}} h_j(x)\,\log T_{ij}\right),
\end{equation}
Here $h_j(x)\in\{0,1,\dots,k\}$ counts the number of neighbors whose observed label equals $j$, hence $\sum_{j\in\mathcal{Y}} h_j(x)=k$. 
If distance weights $w_r$ are used, we replace $h_j(x)$ by the weighted sum $\sum_{r:\,\tilde y_r=j} w_r$. This posterior relies on the conditional independence assumption: given the true label $y$, the observed ratings of neighbors are mutually independent. When $T$ is diagonally dominant (close to $I$), the posterior behavior approaches that of frequency- or average-based voting. If $T = I$ without any smoothing, however, the likelihood degenerates; thus, we apply mild smoothing to $T$ and compute in the log domain to ensure numerical stability. The posterior expectation score is
\begin{equation}
    \hat{s}(x) = \sum_{i=1}^C i \cdot \mathbb{P}(y=i\mid h(x)).
\end{equation}
Finally, the quality reward used in training is given by the piecewise mapping
\begin{equation}
    R_q(x) = 
    \begin{cases}
    1, & \hat{s}(x)\ge 4,\\[2pt]
    0.3, & 3\le \hat{s}(x)<4,\\[2pt]
    0, & \text{otherwise}.
    \end{cases}
\end{equation}
As $T$ becomes diagonally dominant (i.e., $T\approx I$), the posterior concentrates on the most frequent neighbor labels and behaves like smoothed majority/frequency voting.\footnote{We apply mild Laplace smoothing $T\leftarrow (1-\alpha)T+\alpha \mathbf{1}\mathbf{1}^\top/C$ with small $\alpha>0$, followed by row-wise renormalization; computations are carried out in the log domain to avoid underflow.} 
When $T$ departs from $I$, the Bayesian calibration systematically corrects label noise.
\subsection{Details of KNN--Bayes Quality Reward Construction}
\label{sec:detail_knn}

To efficiently approximate LLM ratings during training, we introduce a KNN--Bayes scoring system, which leverages neighborhood information and a score transition matrix to denoise label noise. The construction consists of the following steps:

\paragraph{Offline Asset Construction.}
We collect approximately 100K samples rated by \texttt{ChatGPT-4o-mini} as a reference set.  
We build a KNN index in the embedding space and compute neighbor rating co-occurrence frequencies to estimate both the score transition matrix \(T\) and the label prior \(\mathbf{p}\) offline, prior to model training.

\paragraph{Online Inference and Reward Computation.}
During training, for each generated output \(\hat{Y}\), we retrieve its \(k\) nearest neighbors in the reference set to form a rating histogram \(\mathbf{h}(\hat{Y}) \in \mathbb{R}^C\).  
We then compute the posterior distribution over true labels as
\begin{equation}
    P(y=i \mid \mathbf{h}(\hat{Y})) 
    \propto p_i \cdot \exp\!\left(\sum_{j} h_j(\hat{Y}) \log T_{ij}\right),
    \label{eq:bayes_posterior_appendix}
\end{equation}
and obtain the expected score
\begin{equation}
    \hat{s}(\hat{Y}) = \sum_{i=1}^{C} i \cdot P(y=i \mid \mathbf{h}(\hat{Y})).
    \label{eq:score_expectation_appendix}
\end{equation}

Finally, rewards are assigned using a piecewise mapping:
\begin{equation}
R_q(\hat{Y}) = 
\begin{cases}
1 & \hat{s}(\hat{Y}) \ge 4,\\[2pt]
0.3 & \hat{s}(\hat{Y}) = 3,\\[2pt]
0 & \text{otherwise}.
\end{cases}
\label{eq:quality_reward_appendix}
\end{equation}

\subsection{Consistency Between KNN--Bayes Rating and LLM Scores}
\textbf{Setup.}  
To evaluate the effectiveness of KNN--Bayes in approximating the original LLM scores, we conduct an offline stratified experiment with a reference set and an evaluation set. Given a dataset \(\mathcal{D}\) with LLM-provided scores \(\tilde{y}\), we first split it into a reference set \(\mathcal{B}\) and an evaluation set \(\mathcal{A}\) via stratified sampling to preserve the label distribution of \(\tilde{y}\) across both sets. On \(\mathcal{B}\), we construct a semantic embedding index and estimate the score transition matrix \(T\) and prior distribution \(p\) through neighborhood co-occurrence statistics. For each sample \(x \in \mathcal{A}\), we retrieve its \(k\)-nearest neighbors in \(\mathcal{B}\), obtain the empirical histogram \(h(x)\), and compute the posterior distribution via  
\begin{equation}
\mathbb{P}(y=i \mid h(x)) \propto p_i \prod_{j\in\mathcal{Y}} T_{ij}^{\,h_j(x)}.
\end{equation}
We then calculate the expected score  
\begin{equation}
\hat{s}(x) = \sum_{i\in\mathcal{Y}} i \cdot \mathbb{P}(y=i \mid h(x)).
\end{equation}

\textbf{Metrics.}  
We assess the consistency between KNN--Bayes scores and original LLM scores from two perspectives: distributional divergence and numerical deviation. The distributional divergence is measured by the Jensen--Shannon (JS) divergence:
\begin{equation}
\mathrm{JS}(P,Q) = \tfrac{1}{2} D_{KL}(P \parallel M) + \tfrac{1}{2} D_{KL}(Q \parallel M), 
\quad M = \tfrac{1}{2}(P+Q),
\end{equation}
where $P$ and $Q$ denote the empirical distributions of $\hat{y}(x)$ and $\tilde{y}(x)$, respectively, 
with $\hat{y}(x)=\mathrm{round}(\hat{s}(x))\in\mathcal{Y}$ (alternatively, we bin $\hat{s}(x)$ into the same $C$ categories).
The numerical deviation is quantified using the Mean Absolute Error (MAE):
\begin{equation}
\mathrm{MAE} = \frac{1}{|\mathcal{A}|} \sum_{x \in \mathcal{A}} |\hat{s}(x) - \tilde{y}(x)|.
\end{equation}

\begin{figure}
    \centering
    \includegraphics[width=\linewidth]{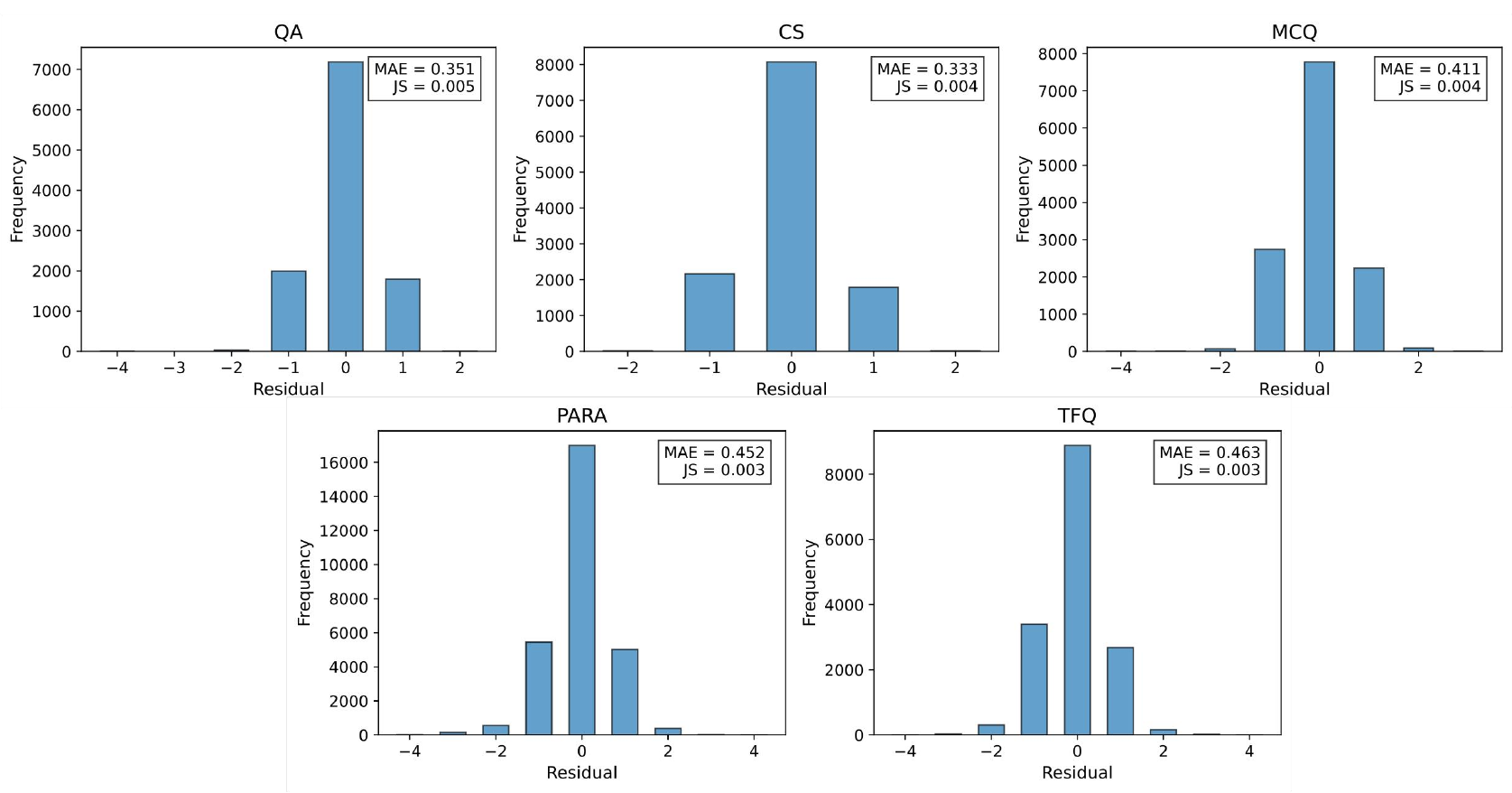}
    \caption{Residual distributions between KNN--Bayes and LLM scores on five datasets. Most residuals concentrate at zero with moderate deviations within $\{-1,1\}$, as shown by MAE and JS metrics.}
    \label{fig:knn_bayes_consistency}
\end{figure}

\textbf{Results.}  
Figure~\ref{fig:knn_bayes_consistency} shows the residual distributions and MAE/JS metrics across all five datasets. Overall, most residuals concentrate at $0$, while some fraction falls within $\{-1,1\}$, suggesting that KNN--Bayes captures the main structure of the original LLM ratings but still exhibits small local deviations. Quantitatively, the JS divergence remains below $0.006$ on all datasets, indicating that the calibrated scores preserve the global distributional shape of the LLM scores with minimal shift. The MAE lies in the range $0.33$–$0.46$, which is moderate compared to the discrete rating scale $\mathcal{Y}=\{0,1,2,3,4,5\}$, reflecting that individual predictions can occasionally deviate by one score level. These findings suggest that while KNN--Bayes provides a low-cost and reasonably accurate approximation for offline evaluation.

\section{Training and Evaluation Details}
\label{sec:setup}
\subsection{Training Details}
We adopt a three-stage training pipeline: (1) \textit{Cold-start full-parameter tuning}, (2) \textit{GRPO reinforcement learning}, and (3) \textit{Evaluation-stage fine-tuning}. All experiments are conducted on 3 H20 GPUs. The key hyperparameters for each stage are summarized below.

\paragraph{Cold-start Training.}
We first perform full-parameter supervised fine-tuning on the initial dataset to provide a strong initialization for later stages. This stage uses a batch size of 128, learning rate $2\times10^{-5}$, and runs for 3 epochs with a maximum sequence length of 2048 tokens.

\paragraph{GRPO Reinforcement Learning.}
The second stage adopts GRPO with multi-dimensional reward signals, including Bayesian KNN-based quality scores, BGE-M3 semantic alignment, and format regularization. We set the rollout batch size to 128, actor global batch size to 16, learning rate $1\times10^{-6}$, KL penalty coefficient $1\times10^{-2}$, and run for 1 epoch with dynamic batching and gradient checkpointing enabled.

\paragraph{Evaluation-stage Fine-tuning.}
Finally, following prior work~\citep{pang2024improving}, we perform lightweight LoRA fine-tuning with a rank size of 64 and a scaling factor of 16. We adopt a batch size of 128, a learning rate of $1\times10^{-4}$, and train for 5 epochs to ensure consistent settings across all evaluation benchmarks.

\begin{table}[h]
\centering
\begin{tabular}{lccc}
\toprule
\textbf{Parameter} & \textbf{Cold-start} & \textbf{GRPO} & \textbf{Eval-tuning} \\
\midrule
Batch size & 128 & 128 (rollout) / 16 (actor) & 128 \\
Learning rate & $2\times10^{-5}$ & $1\times10^{-6}$ & $1\times10^{-4}$ \\
Epochs & 3 & 1 & 5 \\
Max sequence length & 2048 & 2048 & 2048 \\
KL penalty & -- & $1\times10^{-2}$ & -- \\
LoRA rank / scaling & -- & -- & 64 / 16 \\
Dropout rate & 0.1 & 0.1 & 0.1 \\
\bottomrule
\end{tabular}
\caption{Key hyperparameter settings across three training stages.}
\label{tab:training_hyperparams}
\end{table}

\subsection{Evaluation Details}
Following previous work~\citep{pang2024improving}, we evaluate the fine-tuned models on five widely used benchmarks: MMLU~\citep{mmlu}, BBH~\citep{bbh}, GSM8K~\citep{gsm}, TruthfulQA~\citep{lin2021truthfulqa}, and TyDiQA~\citep{clark2020tydi}. For each dataset, we follow standard protocols or common configurations. Specifically, 0-shot settings are used for MMLU; 8-shot in-context examples for GSM8K; 3-shot settings without chain-of-thought for BBH; 6-shot prompts for TruthfulQA; and one in-context example per language for TyDiQA. 

\section{Baseline Details}
\label{app:baseline_details}

We provide detailed descriptions of all baselines considered in the main experiments:

\begin{itemize}[leftmargin=*]
    \item \textbf{Random Selection}: Randomly selects training samples without any filtering.
    \item \textbf{Completion Length}: Uses the total conversation length as a proxy for data quality, assuming longer completions indicate richer information.
    \item \textbf{Perplexity}: Computes perplexity in a zero-shot manner using a pre-trained model; higher perplexity suggests rarer or more complex samples.
    \item \textbf{k-NN}: Measures average distance to the $k$ nearest neighbors in the SentenceBERT embedding space to quantify data rarity.
    \item \textbf{LESS}~\citep{xia2024less}: Scores samples by their influence on a validation set, estimated via gradient-based metrics.
    \item \textbf{AlpaGasus}~\citep{chen2023alpagasus}: Employs an LLM to assign quality ratings, selecting only high-scoring samples.
    \item \textbf{DEITA}~\citep{deita}: Scores samples by both quality and complexity, while iteratively enforcing diversity constraints.
    \item \textbf{DS2}~\citep{pang2024improving}: Selects high-quality and diverse samples by correcting LLM-generated scores via a transition matrix and combining them with long-tail diversity scores.
    \item
    \textbf{EDA}~\citep{wei2019eda}: 
Applies simple text-level perturbations such as synonym replacement, random insertion, random deletion, and word swapping to increase surface-level diversity without altering the original semantic content.
\item
\textbf{Rephrasing} \citep{rephrase}: 
Generates semantically equivalent paraphrases of the original instructions using GPT-4o mini, aiming to modify expression style while preserving meaning to introduce natural linguistic variation.
\item 
\textbf{Back-translation}~\citep{backtrans}: 
Translates each sample into an intermediate language and back to the source language, producing paraphrastic variants that expose the model to diverse lexical and syntactic forms.
    \item \textbf{Full Data}: Uses the entire dataset without any filtering for model fine-tuning.
\end{itemize}

For all rating-based methods (\textit{AlpaGasus}, \textit{DEITA}, and \textit{DS2}), we follow \method and adopt \texttt{ChatGPT-4o-mini} as the rating model for a fair comparison. {\color{red} }

For all data augmentation baselines, the augmentation operations are typically applied to the entire dataset, whereas our other baselines are constructed using 10K training samples. To ensure a fair comparison, we first randomly sample 5K instances from the original 300K data pool and then apply the corresponding augmentation method to these 5K samples, resulting in a total of 10K training examples.

\section{Additional Details of Data Preparation and Experimental Settings}
\label{sec:other_setting}
In our experiments, all 300K samples are annotated with a quality score ranging from 1 to 5. 
Among them, approximately 30K samples with scores $\geq 4$ are treated as high-quality data, 
while roughly 270K samples with scores $< 4$ constitute the low-quality pool. 
Below we provide additional clarifications for the datasets used in Table~\ref{tab:openllm_results_base_llama3.1}.

\paragraph{ORI (High-Quality Data).}
From the 30K high-quality samples, we compute long-tail diversity scores and select the top $N$ instances 
to form the ORI training set. These samples serve as the high-quality-only baseline.

\paragraph{LOW (Low-Quality Data).}
To construct the LOW baseline, we randomly sample $N$ instances from the 270K low-quality pool. 
This setting evaluates the performance of directly using weak signals without any enhancement.

\paragraph{BASE (Qwen-1.5B-Instruct, Zero-shot Generation).}
We directly apply the off-the-shelf Qwen-1.5B-Instruct model (without any training or fine-tuning) 
to fuse inputs from the low-quality pool. For each generated output, we further compute its long-tail diversity score 
and select the top $N$ instances accordingly. The resulting set constitutes the BASE dataset.

\paragraph{MIXUP (LM-Mixup Generated Data).}
The MIXUP dataset is constructed using the following pipeline:
\begin{enumerate}
    \item Sample multiple groups of low-quality inputs from the 270K data pool.
    \item Apply LM-Mixup to each group of $n$ inputs to generate a smaller set of fused, high-quality candidates.
    \item Compute long-tail diversity scores for all mixup-generated outputs.
    \item Select the top 5K samples to form the final MIXUP high-quality dataset.
\end{enumerate}
This process allows LM-Mixup to aggregate and refine weak signals into information-dense, high-quality supervision data.

\section{More Experiment Results}
\label{sec:appen_exper}
\begin{figure}
    \centering
    \includegraphics[width=1.0\linewidth]{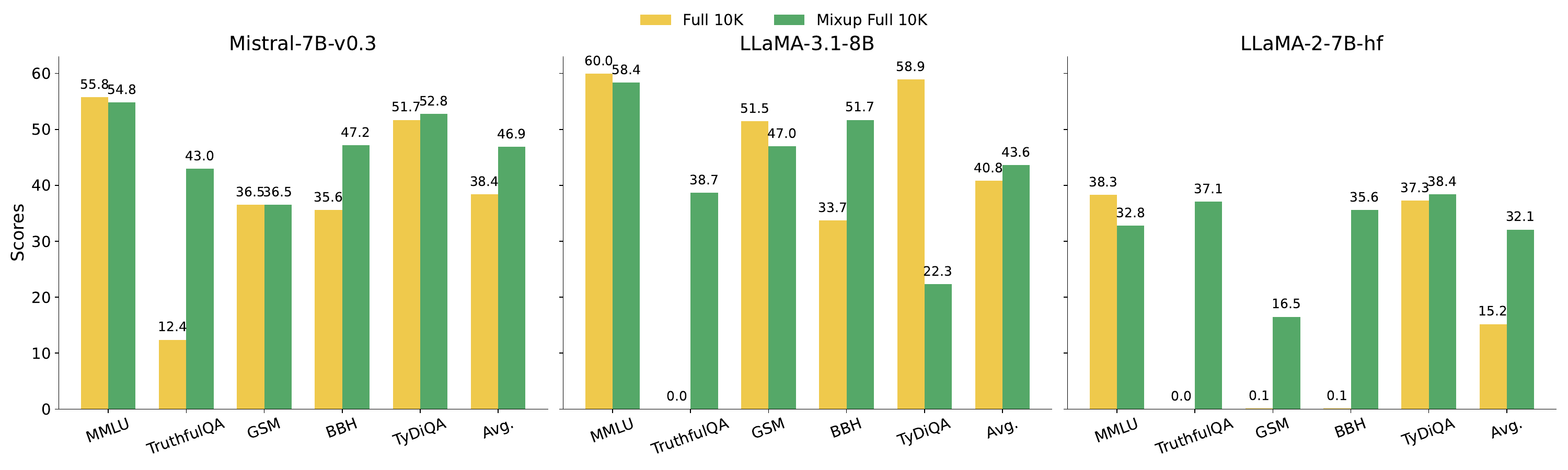}
    \caption{Comparison of model performance on five benchmarks using the full low-quality dataset versus the full mixup dataset (10K samples). Scores are reported for MMLU, TruthfulQA, GSM, BBH, TyDiQA, and the overall average.
}
    \label{fig:full_low_vs_mixup}
\end{figure}
\textbf{Full Low-Quality vs. Full Mixup Data.} To comprehensively evaluate the effectiveness of our approach, we conduct experiments on three representative models—Mistral-7B, LLaMA-3.1-8B, and LLaMA-2-7B-hf—using 10K samples drawn respectively from the raw low-quality dataset and the mixup-enhanced dataset generated via \method. As shown in Figure~\ref{fig:full_low_vs_mixup}, directly fine-tuning on low-quality data leads to inconsistent and often suboptimal performance across most benchmarks. In contrast, the mixup-enhanced data substantially boosts performance on key tasks such as TruthfulQA, BBH, TyDiQA, and the overall average score for all three models. Notably, the improvements are most pronounced on LLaMA-2-7B-hf, where the baseline performance on raw data is particularly low, highlighting the robustness of \method in challenging low-quality settings. These results collectively demonstrate that our method consistently transforms low-quality samples into a valuable resource for instruction tuning, unlocking their potential and significantly narrowing the gap with high-quality data baselines.

\textbf{Additional Results on LLaMA-2-7B-hf and Mistral-7B-v0.3.} We additionally conducted experiments to assess the performance of the OpenLLM leaderboard across different baseline settings using various backbone models, including Mistral-7B-v0.3 and LLaMA-2-7B-hf. Tables \ref{tab:openllm_results_base_mistral} and \ref{tab:openllm_results_base_llama2} report the corresponding results for these two backbones, respectively. Overall, the findings further confirm the effectiveness of our method, demonstrating that with appropriate configurations, it can consistently achieve top-2 performance on the leaderboard.

\begin{figure}
    \centering
    \includegraphics[width=1\linewidth]{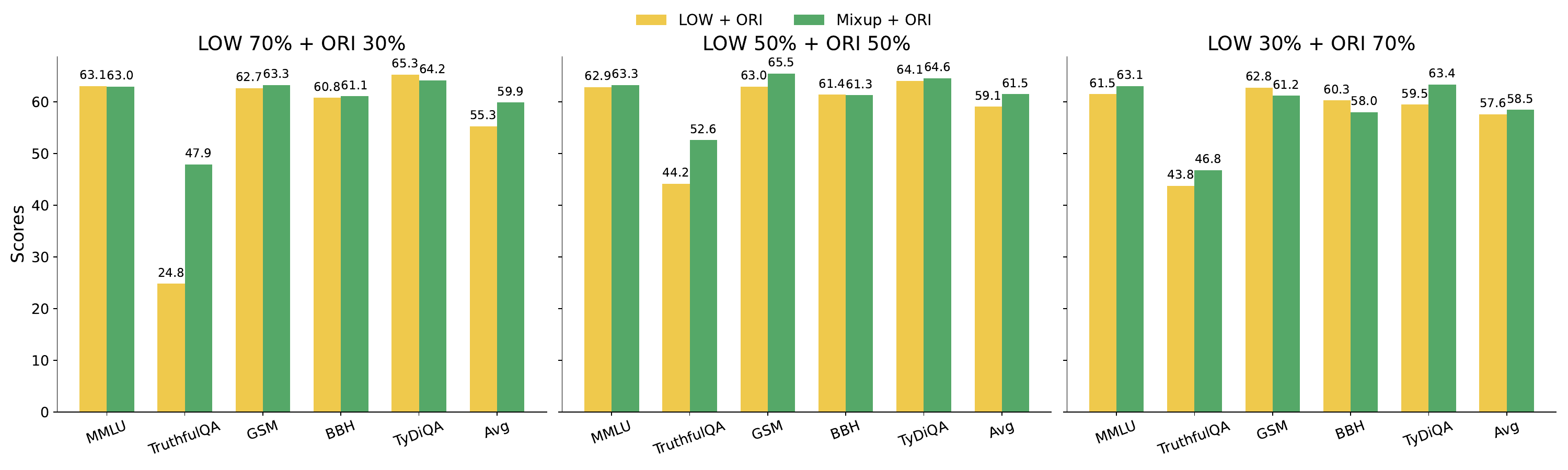}
    \caption{ Comparison between long-tail sampled low-quality data (LOW + ORI) and LM-Mixup (Mixup + ORI) across different data ratios.}
    \label{fig:low_lt}
\end{figure}

\textbf{Analysis of long-tail sampling for low-quality data.}
As shown in Figure~\ref{fig:low_lt}, although replacing random sampling with long-tail sampling for low-quality data brings moderate improvements on certain tasks (e.g., \textit{TruthfulQA}) as well as slight gains in overall average performance, it still consistently underperforms compared to the LM-Mixup method. Under the same data composition settings, LM-Mixup achieves more substantial and stable improvements across multiple benchmarks, particularly on \textit{TruthfulQA} and \textit{GSM}. This suggests that merely refining the sampling strategy of low-quality data is insufficient to address its inherent issues, such as incomplete information and fragmented expression. In contrast, LM-Mixup effectively integrates multiple low-quality inputs into more coherent and informative training signals, leading to more comprehensive performance gains. These results further demonstrate the superiority of our method in improving data effectiveness.

\begin{table}[h]
    \centering
    \resizebox{\linewidth}{!}{
    \begin{tabular}{l|ccccccc}
    \toprule
    \multirow{2}{*}{\textbf{Models}} & \multicolumn{1}{c}{\textbf{MMLU}} & \multicolumn{1}{c}{\textbf{TruthfulQA}} & \multicolumn{1}{c}{\textbf{GSM}} & \multicolumn{1}{c}{\textbf{BBH}} & \multicolumn{1}{c}{\textbf{TydiQA}} & \multicolumn{1}{c}{\textbf{Average}} \\
    & (\textbf{factuality}) & (\textbf{truthfulness}) & (\textbf{reasoning}) & (\textbf{reasoning})  & (\textbf{multilinguality})  & \\
    \midrule
    \textsc{Vanilla base model*} & 59.7 & 30.2 & 38.0 & 49.6 & 54.9 & 46.5 \\
    \textsc{Completion length*}   & 58.9 & 34.4 & 42.5 & 53.1 & 59.6 & 49.7\\
    \textsc{Perplexity*}   & 59.8 & 40.3 & 36.0 & 48.9 & 57.4 & 48.5 \\
    \textsc{$k$-NN-10*}   & 58.3 & 41.7 & 43.5 & 54.1 & 53.4 & 50.2\\
    \textsc{Random Selection*} & 59.4 & 36.7 & 41.8 & 54.2  & 54.0 & 49.3 \\
    \textsc{LESS*}    & 59.5 & 34.8 & 42.0 & 54.5 & 57.5 & 49.7\\
    \textsc{Full data (300K)*} & 60.0    &    43.5 & 43.5 & 52.5  &  53.4   &      50.6 \\
    \midrule 
    \textsc{AlpaGasus*}    & 60.5     &   36.7  &41.0 & 55.1   & 57.3       &  50.1 \\
    \textsc{Deita*}    & 60.1   &     35.6 & 40.5  &55.1  &  56.0    &     49.5\\
    \textsc{DS2 w/o curation*}  &60.1    &    35.9 & 48.5  &54.2  &  58.9     &    51.5 \\
    \textsc{DS2*}  &59.9    &    37.9 & 47.5  &55.6  &  59.3     &    \textbf{52.0} \\
    \midrule 
    \textsc{Mixup 70\% + Ori 30\%} & 58.5 & 42.7 & 46.0 & 53.2 & 52.9 & 50.7 \\
\textsc{Mixup 50\% + Ori 50\%} & 57.0 & 43.0 & 47.0 & 54.0 & 52.6 & 50.7 \\
\textsc{Mixup 30\% + Ori 70\%} & 56.0 & 45.3 & 51.5 & 54.0 & 52.1 & \underline{51.8} \\
    \bottomrule
    \end{tabular}
    }
    \caption{Results on the OpenLLM leaderboard using \texttt{Mistral-7B-v0.3} as the base model. The top-performing scores are shown in \textbf{bold}, while the second-best scores are marked with \underline{underlines}. * indicates that the values are sourced from \cite{pang2024improving}.}
    \label{tab:openllm_results_base_mistral}
\end{table}

\begin{table}[h]
    \centering
    \resizebox{\linewidth}{!}{
    \begin{tabular}{l|ccccccc}
    \toprule
    \multirow{2}{*}{\textbf{Model}} & \multicolumn{1}{c}{\textbf{MMLU}} & \multicolumn{1}{c}{\textbf{TruthfulQA}} & \multicolumn{1}{c}{\textbf{GSM}} & \multicolumn{1}{c}{\textbf{BBH}} & \multicolumn{1}{c}{\textbf{TydiQA}} & \multicolumn{1}{c}{\textbf{Average}} \\
    & (\textbf{factuality}) & (\textbf{truthfulness}) & (\textbf{reasoning}) & (\textbf{reasoning})  & (\textbf{multilinguality})  & \\
    \midrule
    \textsc{Vanilla LLaMa-2-7B*} & 41.9 & 28.4 &  6.0 & 38.3 & 35.7 & 30.1 \\
    \textsc{Completion Length*} & 42.4 & 36.4 & 1.5 & 36.8 & 33.9 & 30.2\\
    \textsc{Perplexity*} & 45.0 & 41.5 & 12.0 & 31.7 & 39.5 &  33.9 \\
    \textsc{$k$-NN-10*} & 38.2 & 40.8 & 15.0 & 36.0 & 43.8 & 34.8\\
   \textsc{Random Selection*} & 44.7  & 41.8  & 14.0  & 37.9 & 40.8  & 35.8 \\
    \textsc{LESS} & 44.3   &     38.2 & 18.0&  35.2 &   46.3         &36.4\\
    \textsc{Full data (300K)*} & 50.1 & 36.2 &  16.5 & 40.5 & 46.7 & 38.0 \\
    \midrule 
    \textsc{AlpaGasus*} &  45.3    &     41.0 &  14.5  & 37.0  &   45.3    &     \underline{36.6} \\
    \textsc{Deita*} & 45.2   &     44.7 & 13.5 & 35.6  &  43.4    &     36.5 \\
    \textsc{DS2 w/o curation*} & 42.0  &      39.5 & 15.0 & 38.1    &46.1     &    36.1 \\
\textsc{DS2*}  & 40.2   &     43.8 & 13.5 & 38.9&    46.5   &      \underline{36.6} \\ %
    \midrule 
    \textsc{Mixup 70\% + Ori 30\%} & 39.5 & 42.5 & 17.0 & 38.6 & 42.5 & 36.0 \\
\textsc{Mixup 50\% + Ori 50\%} & 39.5 & 44.0 & 18.0 & 38 & 42.5 & 36.4 \\
\textsc{Mixup 30\% + Ori 70\%} & 39.0 & 45.3 & 18.0 & 38.0 & 43.5 & \textbf{36.8} \\
    \bottomrule
    \end{tabular}
    }
    \caption{Results on the OpenLLM leaderboard using \texttt{LLaMA-2-7B-hf} as the base model. The top-performing scores are shown in \textbf{bold}, while the second-best scores are marked with \underline{underlines}. * indicates that the values are sourced from \cite{pang2024improving}.}
    \label{tab:openllm_results_base_llama2}
\end{table}

\section{Additional Dataset Statistics about \dataset}
\begin{figure}
    \centering
    \includegraphics[width=\linewidth]{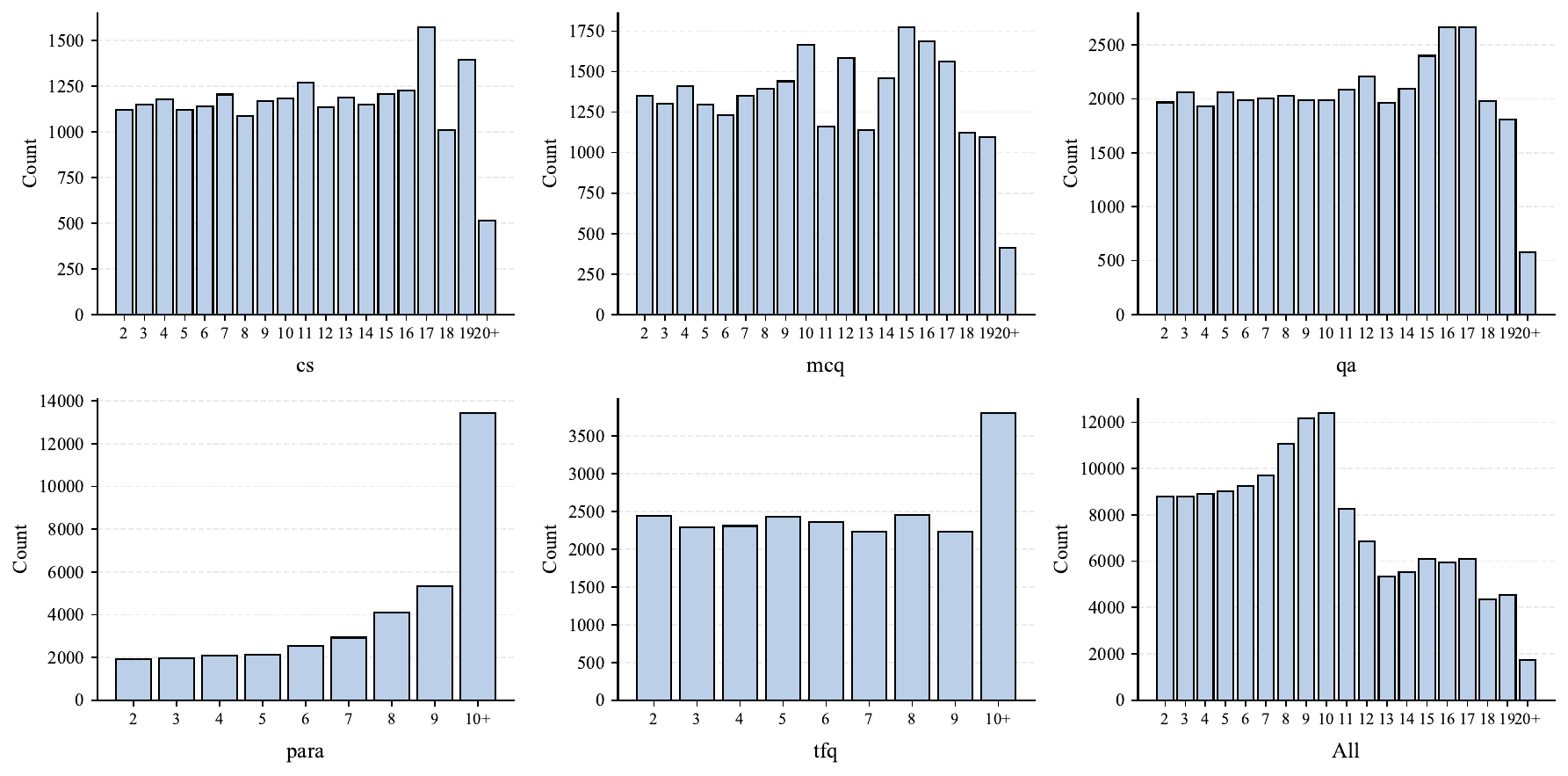}
    \caption{Distribution of the number of low-quality samples derived from each high-quality sample across different task categories.}
    \label{fig:split_count}
\end{figure}

Figure~\ref{fig:split_count} reports the distribution of the number of low-quality variants constructed for each high-quality sample. Most samples are paired with multiple degraded variants, enabling the model to learn hierarchical mappings from noisy or incomplete inputs to high-quality outputs.

Table~\ref{tab:mixup_category_breakdown} provides a detailed breakdown of the entire dataset across five task types (QA, MCQ, CS, TFQ, Paragraph) and three data variants (Normal, Cross-Topic, Noisy). We observe a balanced distribution across task types, with QA and Paragraph slightly larger in size, ensuring diverse coverage for training and evaluation.

\section{Prompt Template}
\label{sec:prompt}
The prompt template below illustrates how we use ChatGPT-4o-mini to generate high-quality data, with the following notes for different task types:
For qa, let Instruction be the question and Output be the answer.
For mcq, include options inside Instruction and provide the correct choice and a brief rationale in Output.
For cs, tfq, or paragraph styles, keep Instruction as the task prompt and Output as the targeted response.

\begin{tcolorbox}[colback=mylightgray, colframe=mygray, title=Prompt Template for High Quality Data Generation]\label{box:gen_template}
\noindent
You are a knowledgeable assistant tasked with producing \textbf{exceptionally high-quality \{task\_type\}} instances that will later be rated on four axes: \emph{Rarity}, \emph{Complexity}, \emph{Informativeness}, and \emph{Overall}.

\medskip
\noindent\textbf{Scoring Criteria}
\begin{itemize}
  \item \textbf{Rarity (1--10):} Cover non-obvious, less-quoted aspects; avoid commonplace trivia.
  \item \textbf{Complexity (1--10):} Require synthesis of multiple facts, causal/temporal links, or non-trivial reasoning; avoid single-sentence lookups.
  \item \textbf{Informativeness (1--10):} Deliver dense, relevant, non-trivial content—even if concise; add value beyond superficial recall.
  \item \textbf{Overall (1--10):} Aggregate impression; aim for the top tier when justified.
\end{itemize}

\medskip
\noindent\textbf{Requirements}
\begin{enumerate}
  \item Generate \textbf{3--4} high-quality \textbf{\{task\_type\}} instances based on the passage below.
  \item For each instance, explicitly maximize the four criteria above (Rarity, Complexity, Informativeness, Overall).
  \item You may quote or paraphrase the passage, but \emph{weave} information to show reasoning and uniqueness; avoid verbatim copying when unnecessary.
  \item Length is flexible—prioritize informativeness and reasoning over verbosity.
  \item Use \textbf{plain text only} in the following exact format (no markdown):
\end{enumerate}

\small
\begin{verbatim}
Instruction: <instruction 1>
Output: <output 1>

Instruction: <instruction 2>
Output: <output 2>

Instruction: <instruction 3>
Output: <output 3>
\end{verbatim}
\normalsize
\medskip

\noindent\textbf{Passage}
\begin{verbatim}
{passage}
\end{verbatim}
\end{tcolorbox}

The prompt template below illustrates how we use ChatGPT-4o-mini to generate low quality data, with the following notes for different task types:
For qa, let Instruction be the question and Output be the answer.
For mcq, include options inside Instruction and provide the correct choice with minimal explanation in Output.
For cs, tfq, or paragraph styles, keep Instruction as the task prompt and Output as the response, ensuring only moderate relevance and detail. The number n is determined by a random value.

\begin{tcolorbox}[colback=mylightgray, colframe=mygray, title=Prompt Template for Low Quality Data Generation]\label{box:low_quality_template}
\noindent
You are a moderately skilled assistant tasked with producing \textbf{\{n\} low-quality \{task\_type\}} instances derived from the given original data. Each generated instance should deliberately reflect \emph{average quality}, aiming for scores between \textbf{3--6 (on a 10-point scale)} on four evaluation axes: \emph{Rarity}, \emph{Complexity}, \emph{Informativeness}, and \emph{Overall}.

\medskip
\noindent\textbf{Scoring Criteria}
\begin{itemize}
  \item \textbf{Rarity (3--6):} Cover reasonably common aspects; avoid both overly trivial and highly novel content.
  \item \textbf{Complexity (3--6):} Allow some light inference but avoid deep reasoning or multi-step logic.
  \item \textbf{Informativeness (3--6):} Ensure answers are mostly correct but lack nuance, depth, or precision.
  \item \textbf{Overall (3--6):} The overall impression should feel average, slightly useful, somewhat generic, and unpolished.
\end{itemize}

\medskip
\noindent\textbf{Hard Constraints}
\begin{enumerate}
  \item \textbf{Same Topic}: All \{task\_type\} instances must stay on the identical topic as the original.
  \item \textbf{Explicit Subject}: The main subject or event (e.g., \emph{the execution of Turner and McDaniel}) \emph{must be stated verbatim} in every question and answer. Avoid vague pronouns unless the noun is immediately repeated.
  \item \textbf{Self-contained}: Each \{task\_type\} must be understandable in isolation; assume no external context.
  \item \textbf{No Off-topic Content}: Do not introduce unrelated domains or shift the factual focus.
\end{enumerate}

\medskip
\noindent\textbf{Output Requirements}
\begin{enumerate}
  \item Generate \textbf{3--4} \{task\_type\} instances based on the original data.
  \item Maintain moderate quality (scores 3--6) on all four evaluation axes.
  \item Use \textbf{plain text only} in the exact format below (no markdown):
\end{enumerate}

\small
\begin{verbatim}
Instruction: <instruction 1>
Output: <output 1>

Instruction: <instruction 2>
Output: <output 2>

Instruction: <instruction 3>
Output: <output 3>
\end{verbatim}
\normalsize
\medskip

\noindent\textbf{Original Data}
\begin{verbatim}
{orig}
\end{verbatim}
\end{tcolorbox}

The prompt template below illustrates how we use ChatGPT-4o-mini to perform data fusion across different task types, with the following notes:
For qa, merge two question–answer pairs into a single, coherent question with a unified answer.
For mcq, combine two multiple-choice questions into one integrated question, providing a single correct option with a concise explanation.
For cs, tfq, or paragraph tasks, merge the content of both instances into a single prompt–response pair, ensuring the output reflects a natural synthesis of the original information while maintaining moderate length and relevance.

\begin{tcolorbox}[colback=mylightgray, colframe=mygray, title=Prompt Template for Data Fusion across \{task\_type\}]\label{box:data_fusion_template}
\noindent
You are a data fusion expert tasked with merging two \{task\_type\} instances into a single, coherent, and high-quality \{task\_type\} instance. The goal is to synthesize both original samples into one unified output that naturally connects the information from both inputs while maintaining high quality.

\medskip
\noindent\textbf{Fusion Requirements}
\begin{enumerate}
  \item Merge the two original \{task\_type\} samples into \textbf{one concise, integrated instance}.
  \item The unified output must address both original inputs comprehensively while avoiding redundancy or contradiction.
  \item The fusion should capture \emph{subtle conceptual links}, rather than simply stacking facts together.
  \item Ensure the final output meets the following quality criteria:
  \begin{itemize}
      \item \textbf{Rarity:} Avoid overly common or trivial facts; focus on non-obvious insights.
      \item \textbf{Complexity:} Encourage nuanced reasoning or implicit connections.
      \item \textbf{Informativeness:} Maximize factual density and relevance.
      \item \textbf{Overall Quality:} Aim for the top tier across all above dimensions.
  \end{itemize}
\end{enumerate}

\medskip
\noindent\textbf{Output Format}
\begin{verbatim}
Instruction: <your merged instruction>
Output: <your merged output>
\end{verbatim}

\medskip
\noindent\textbf{Original Instances}
\begin{verbatim}
Instance-1:
{text1}

Instance-2:
{text2}
\end{verbatim}
\end{tcolorbox}

Following previous work~\citep{pang2024improving}, we use the same template for LLM Rating:
\begin{tcolorbox}[colback=mylightgray, colframe=mygray, title=Prompt Template for LLM Rating]\label{box:prompt_template}

\noindent
As a data quality estimator, your task is to assess the quality of the data sample based on the criteria: Rarity, Complexity, and Informativeness. Please rate the sample on a scale from 1 to 10 for each criterion, and return an overall rating on a scale from 1 to 10, where a higher score indicates a higher level of quality. Ensure that the ratings are not overly concentrated around a specific score. If multiple samples have similar qualities, consider spreading the scores more evenly to reflect subtle differences.  \\

Please carefully evaluate the following data sample and return the integral evaluation scores using the JSON format:
    \small 
    \begin{verbatim}
    {"Rarity": <number, 1-10>,
        "Complexity": <number, 1-10>,
        "Informativeness": <number, 1-10>,
        "Overall rating": <number, 1-10>}
    \end{verbatim}
    \normalsize 
Instruction: [Instruction] \\
Input: [Input] \\ 
Response: [Response] 
\end{tcolorbox}

\section{Case Study}
\label{sec:case_study}
Tables~\ref{tab:qa_case}, \ref{tab:cs_mixup_case}, \ref{tab:tfq_mixup_case}, \ref{tab:mcq_mixup_case}, and \ref{tab:para_case} present representative mixup cases across QA, CS, TFQ, MCQ, and paragraph tasks, illustrating how multiple fragmented inputs are integrated into coherent, high-quality outputs.
\definecolor{darkblue}{RGB}{70,130,180}  
\definecolor{lightblue}{RGB}{173,216,230}  
\definecolor{paleblue}{RGB}{220,240,255}  
\definecolor{verylightblue}{RGB}{240,248,255} 
\begin{table}[htbp]
\centering
\resizebox{1\linewidth}{!}{
\begin{tabular}{@{}p{14cm}@{}}
\toprule
\rowcolor{lightblue}
\textbf{Input (Multiple QA Pairs)}\\ 
\midrule
\rowcolor{verylightblue!98} 
Q: What are the financial implications Hardy discussed with the AAHA and Saskatchewan Amateur Hockey Association regarding the discussions on player development?  

A: Hardy's discussions highlighted that financial implications include a strain on resources, showing the need for more investment in player retention to prevent losing talent to professional leagues.

Q: How might the discussions with the AAHA and Saskatchewan Amateur Hockey Association on player development affect the competitiveness of amateur hockey?  

A: The discussions suggest that amateur hockey may face challenges to its competitiveness if it cannot adapt to the pressures from professional leagues, potentially leading to a decline in player quality.

Q: What strategic changes were suggested in Hardy's discussions with the AAHA and Saskatchewan Amateur Hockey Association regarding player development?  

A: Hardy's discussions prompted the idea of re-evaluating investment strategies in player development to ensure that amateur hockey remains appealing and retains its talent pool amidst increasing professional competition.

Q: What broader challenges were mentioned by Hardy concerning the discussions on player development?  

A: Hardy mentioned that broader challenges include maintaining the integrity of amateur hockey and ensuring grassroots involvement, which are at risk due to the rising influence of professional leagues. \\
\midrule
\rowcolor{darkblue}
\textcolor{white}{\textbf{Output (Merged Answer)}} \\ 
\midrule
\rowcolor{verylightblue!98} 
Q: Analyze how Hardy’s post-WWII discussions with amateur hockey associations reflected emerging tensions in Canadian sports culture, particularly regarding financial constraints, competitiveness, and societal identity shifts.

A: In examining Hardy's dialogue with prominent figures representing national governing bodies post-WWII, there emerges a profound intersectionality of personal ambition juxtaposed against systemic constraints faced by athletes transitioning towards professional careers. His candid remarks reveal underlying concerns about the sustainability of amateurism given escalating demands placed upon young talents competing internationally. Such sentiments reflect wider cultural shifts occurring globally wherein individuals sought greater recognition beyond localized boundaries—a trend mirrored in other sporting disciplines experiencing rapid modernization. Furthermore, Hardy's reflections hint at deeper socio-political ramifications tied to identity formation linked to sport participation, suggesting that individual aspirations intersected critically with collective narratives shaped by historical events affecting community cohesion. Overall, these discussions underscore significant transformations happening simultaneously in society and athletics, revealing complexities inherent in balancing tradition with contemporary realities facing aspiring amateurs navigating pathways toward professional success. \\
\bottomrule
\end{tabular}}
\caption{QA pairs mixup example.}
\label{tab:qa_case}
\end{table}

\begin{table}[htbp]
\centering
\resizebox{1\linewidth}{!}{
\begin{tabular}{@{}p{14cm}@{}}
\toprule
\rowcolor{lightblue}
\textbf{Input (Multiple CS Pairs)} \\
\midrule

\rowcolor{verylightblue!98}
C: Historical evolution of electromagnetic theory 

S: The historical evolution of electromagnetic theory involves key figures like Maxwell who explored the aether theory, which proposed that forces act through a medium.

C: Philosophical implications of electromagnetic theory 

S: The philosophical implications of electromagnetic theory reflect a shift in understanding forces, moving from the aether concept to recognizing the nature of fields and potentials in physics.

C: Impact of electromagnetic theory on engineering 

S: The impact of electromagnetic theory on engineering principles is substantial, as it provides a framework for understanding different physical properties compared to concepts like potential energy.

C: The role of experiments in electromagnetic theory 

S: The role of experiments in electromagnetic theory, particularly the Michelson–Morley experiment, played a crucial part in dismissing the aether theory and reshaping modern physics. \\
\midrule
\rowcolor{darkblue}
\textcolor{white}{\textbf{Output (Merged Answer)} }
\\
\midrule
\rowcolor{verylightblue!98}
C: Synthesis of Physics Concepts 

S: The exploration of electromagnetic phenomena reflects significant advancements in classical mechanics, wherein the transition from Newtonian dynamics towards relativistic formulations necessitated a paradigmatic change akin to shifting away from traditional force-based models toward field-theoretic approaches rooted in differential forms. While early attempts at explaining electrostatics involved complex interactions mediated via hypothetical entities known as electric fluids, later thinkers sought simpler explanations grounded in mathematical constructs involving vector calculus. Notably, the dismissal of the aether hypothesis postulated by Fresnel was pivotal, leading physicists to embrace alternative interpretations supported by extensive experimental validation, including notable outcomes observed in the Michelson–Morley interferometer experiment. These evolving understandings illustrate profound intersections between philosophy and science, revealing deeper layers of reality beyond conventional boundaries. \\
\bottomrule
\end{tabular}}
\caption{CS pairs mixup example.}
\label{tab:cs_mixup_case}
\end{table}

\begin{table}[htbp]
\centering
\resizebox{1\linewidth}{!}{
\begin{tabular}{@{}p{14cm}@{}}
\toprule
\rowcolor{lightblue}
\textbf{Input (Multiple TFQ Pairs)} \\
\midrule
\rowcolor{verylightblue!98}
Joanne Peason's view of Pryce's essay does not show an important moment in the debate about Christianity's origins in Britain. 

Answer: False

Joanne Peason believes that Pryce's essay fails to effectively challenge established legends. 

Answer: False

Pryce's work is considered by Joanne Peason to have no relevance to discussions of Welsh identity and Christianity. 

Answer: False \\
\midrule
\rowcolor{darkblue}
\textcolor{white}{\textbf{Output (Merged Answer)}}
 \\
\midrule
\rowcolor{verylightblue!98}
Joanne Pearson's assertion that Pryce's essay represents "a significant point" in challenging prevailing myths suggests she recognizes the importance of questioning traditional accounts of early Christianization processes in Britain, especially relating to Welsh culture. Her belief indicates a shift towards more nuanced interpretations where local identities were intertwined with religious conversion stories during periods like Roman rule. Such skepticism contributes significantly to ongoing debates over national histories shaped by colonial influences, suggesting that Pryce's exploration may serve dual purposes: critiquing existing narratives and fostering contemporary dialogues about indigenous rights and heritage preservation amidst modernity. 

Answer: False \\
\bottomrule
\end{tabular}}
\caption{TFQ pairs mixup example.}
\label{tab:tfq_mixup_case}
\end{table}

\begin{table}[htbp]
\centering
\resizebox{1\linewidth}{!}{
\begin{tabular}{@{}p{14cm}@{}}
\toprule
\rowcolor{lightblue}
\textbf{Input (Multiple MCQ Pairs)} \\
\midrule
\rowcolor{verylightblue!98}
Q: What is a possible effect of contestants giving “Bonus” points to each other throughout the show? 

A) It might lead to an increased sense of competition among contestants. 

B) Contestants may use Bonus points to create alliances, improving team dynamics. 

C) The Bonus points could make contestants less motivated to help one another. 

D) Contestants might ignore Bonus points and only care about the judges’ scores. 

Answer: B

Q: How do Bonus points awarded by contestants impact their relationships? 

A) They may cause contestants to form rivalries that hurt their performance. 

B) Contestants could leverage Bonus points to build alliances and boost collaboration. 

C) The use of Bonus points may lower motivation among contestants. 

D) Contestants generally overlook Bonus points, focusing only on judges' feedback. 

Answer: B

Q: In the context of the show, how might the Bonus points affect contestant interactions? 

A) They could lead to increased competition and less teamwork. 

B) Contestants might use Bonus points to create positive alliances, enhancing performance. 

C) The Bonus points system could demotivate contestants from giving constructive feedback. 

D) Contestants may disregard Bonus points, concentrating on judges' assessments instead. 

Answer: B

Q: What impact do Bonus points have on how contestants relate to one another? 

A) They might foster rivalries that detract from overall performance. 

B) Contestants may utilize Bonus points to form alliances, which can enhance their teamwork. 

C) The Bonus points system could lead to a lack of motivation among the contestants. 

D) Contestants often pay little attention to Bonus points, focusing more on judges’ scores. 

Answer: B \\
\midrule
\rowcolor{darkblue}
\textcolor{white}{\textbf{Output (Merged MCQ Instance)}}
 \\
\midrule
\rowcolor{verylightblue!98}
Q: Analyzing the psychological dimensions involved in the game structure reveals insights into how power operates differently than traditional authority figures. Consider the role of contestants awarding "Bonus" points to each other over time. Discuss the broader implications of this mechanism on perceptions of fairness and trustworthiness among competitors. Which statement best captures the dual nature of this dynamic? 

A) Awarding Bonus points fosters genuine camaraderie among contestants, leading to improved performances due to mutual support rather than fear-based compliance typical of hierarchical setups. 

B) By allowing contestants to give Bonus points freely, the hierarchy becomes irrelevant, suggesting that true leadership emerges solely from meritocratic principles regardless of initial status differences. 

C) The ability to manipulate others through Bonus points creates distrust among teammates, undermining collaborative efforts necessary for success, thereby reinforcing existing hierarchies despite attempts to level playing fields. 

D) Giving Bonus points reflects a shift towards egalitarianism, enabling weaker members to challenge dominant personalities directly, ultimately reshaping societal norms beyond just entertainment purposes. 

Answer: C \\
\bottomrule
\end{tabular}}
\caption{MCQ pairs mixup exmaple.}
\label{tab:mcq_mixup_case}
\end{table}

\begin{table}[htbp]
\centering
\resizebox{1\linewidth}{!}{
\begin{tabular}{@{}p{14cm}@{}}
\toprule
\rowcolor{lightblue}
\textbf{Input (Four Separate Paragraphs)} \\
\midrule
\rowcolor{verylightblue!98}
1. The Industrial Revolution transformed urban landscapes across Europe, bringing rapid technological advancements but also leading to overcrowding and unsanitary living conditions in many cities. 

2. During the same period, public health reforms emerged as governments recognized the need for clean water systems, waste management, and regulations to combat disease outbreaks. 

3. Architects and city planners began integrating green spaces and wider streets into urban designs, seeking to improve both public health and aesthetic appeal. 

4. These efforts laid the groundwork for modern urban planning, where functionality, sustainability, and livability became central considerations in shaping city environments. \\
\midrule
\rowcolor{darkblue}\textcolor{white}{\textbf{Output (Merged High-Quality Paragraph)}}
 \\
\midrule
\rowcolor{verylightblue!98}
The Industrial Revolution reshaped European cities through rapid industrialization, introducing new technologies but also creating overcrowded and unhealthy living conditions. As governments grappled with recurring disease outbreaks, public health reforms prioritized clean water systems, sanitation, and housing regulations. Urban planners responded by incorporating green spaces, broader streets, and organized infrastructure, aiming to balance industrial growth with livable environments. These combined efforts forged the foundations of modern urban planning, where health, sustainability, and functionality guide the evolution of contemporary cities. \\
\bottomrule
\end{tabular}}
\caption{Paragraphs mixup example.}
\label{tab:para_case}
\end{table}

\newpage
\section{Dataset Statistics}
\label{ap:data_stat}

\begin{table}[h]
\centering
\fontsize{8pt}{10pt}\selectfont
\begin{tabular}{l@{\hskip 1.5mm}rr}
\toprule
\rowcolor{gray!15}
\textbf{Category} & \textbf{Samples} & \textbf{Ratio (\%)} \\
\midrule

\rowcolor{CS}
\textbf{I. cs} & \textbf{22,012} & \textbf{15.2} \\
\hspace{8pt}\textbullet\hspace{4pt}Noisy & 1,100 & 0.8 \\
\hspace{8pt}\textbullet\hspace{4pt}Cross-Topic & 4,852 & 3.3 \\
\hspace{8pt}\textbullet\hspace{4pt}Normal & 16,060 & 11.1 \\

\rowcolor{MCQ}
\textbf{II. mcq} & \textbf{25,437} & \textbf{17.6} \\
\hspace{8pt}\textbullet\hspace{4pt}Noisy & 1,272 & 0.9 \\
\hspace{8pt}\textbullet\hspace{4pt}Cross-Topic & 4,714 & 3.3 \\
\hspace{8pt}\textbullet\hspace{4pt}Normal & 19,451 & 13.4 \\

\rowcolor{QA}
\textbf{III. qa} & \textbf{38,455} & \textbf{26.6} \\
\hspace{8pt}\textbullet\hspace{4pt}Noisy & 1,922 & 1.3 \\
\hspace{8pt}\textbullet\hspace{4pt}Cross-Topic & 7,405 & 5.1 \\
\hspace{8pt}\textbullet\hspace{4pt}Normal & 29,128 & 20.1 \\

\rowcolor{PARA}
\textbf{IV. para} & \textbf{36,423} & \textbf{25.1} \\
\hspace{8pt}\textbullet\hspace{4pt}Noisy & 1,821 & 1.3 \\
\hspace{8pt}\textbullet\hspace{4pt}Cross-Topic & 8,261 & 5.7 \\
\hspace{8pt}\textbullet\hspace{4pt}Normal & 26,341 & 18.2 \\

\rowcolor{TFQ}
\textbf{V. tfq} & \textbf{22,557} & \textbf{15.6} \\
\hspace{8pt}\textbullet\hspace{4pt}Noisy & 1,128 & 0.8 \\
\hspace{8pt}\textbullet\hspace{4pt}Cross-Topic & 4,154 & 2.9 \\
\hspace{8pt}\textbullet\hspace{4pt}Normal & 17,275 & 11.9 \\

\rowcolor{blue!10}
\textbf{All} & \textbf{144,884} & \textbf{100.0} \\
\bottomrule
\end{tabular}
\caption{Dataset statistics across five task categories. For each category, we report the total number of samples (train+test) and their breakdown into noisy, cross-Topic, and normal subsets. Ratios indicate the percentage of the full dataset.}
\label{tab:mixup_category_breakdown}
\end{table}

\end{document}